\documentclass[11pt,doublespace]{article}
\usepackage{cite}
\usepackage{setspace}

\usepackage[cmex10]{amsmath}
\usepackage{amssymb,tabularx}
\usepackage{algorithmic}
\usepackage{array}

  \usepackage[]{graphicx}

\usepackage{fixltx2e}
\usepackage{dblfloatfix}

  \usepackage[nomarkers]{endfloat}

\begin{document}
%
\title{A variational Bayes framework for sparse adaptive estimation}
\author{Konstantinos E.Themelis, Athanasios~A.~Rontogiannis, \\ and Konstantinos D. Koutroumbas%
\thanks{K. E. Themelis, A. A. Rontogiannis and K. D. Koutroumbas are with the Institute for Astronomy, Astrophysics, Space Applications and Remote Sensing (IAASARS), National Observatory of Athens, I. Metaxa \& Vas. Pavlou str., GR-15236, Penteli, Greece (e-mail: themelis@noa.gr; tronto@noa.gr; koutroum@noa.gr).}
}




\maketitle

\begin{abstract}
Recently, a number of mostly $\ell_1$-norm regularized least squares type deterministic algorithms have been proposed to address the problem of \emph{sparse} adaptive signal estimation and system identification. From a Bayesian perspective, this task is equivalent to maximum a posteriori probability estimation under a sparsity promoting heavy-tailed prior for the parameters of interest. Following a different approach, this paper develops a unifying framework of sparse \emph{variational Bayes} algorithms that employ heavy-tailed priors in conjugate hierarchical form to facilitate posterior inference. The resulting fully automated variational schemes are first presented in a batch iterative form. Then it is shown that by properly exploiting the structure of the batch estimation task, new sparse adaptive variational Bayes algorithms can be derived, which have the ability to impose and track sparsity during real-time processing in a time-varying environment. The most important feature of the proposed algorithms 
is 
that they 
completely eliminate the need for computationally costly parameter fine-tuning, a necessary ingredient of sparse adaptive deterministic algorithms. 
Extensive simulation results are provided to demonstrate the effectiveness of the new sparse variational Bayes algorithms against state-of-the-art deterministic techniques for adaptive channel estimation. The results show that the proposed algorithms are numerically robust and exhibit in general superior estimation performance compared to their deterministic counterparts. 
\end{abstract}


%


\section{Introduction}

Adaptive estimation of time-varying signals and systems is a research field that has attracted tremendous attention in the statistical signal processing literature, has triggered extensive research, and has had a great impact in a plethora of applications \cite{2001_haykin, 2008_sayed}. A large number of adaptive estimation techniques have been developed and analyzed during the past decades, which have the ability to process streaming data and provide real-time estimates of the parameters of interest in an online fashion. It has long ago been recognized that apart from being time-varying, most signals and systems, both natural and man-made, also admit a parsimonious or so-called \emph{sparse} representation in a certain domain. This fact has nowadays sparkled new interest in the area of adaptive estimation, as the recent advances and tools developed in the compressive sensing (CS) field \cite{2008_candes,2006_donoho}, provide the means to effectively exploit {\it sparsity} in a time-varying 
environment. 
It has been anticipated that by suitably exploiting system sparsity, significant improvements in convergence rate and estimation performance of adaptive techniques could be achieved.

It is not surprising that the majority of sparsity aware adaptive estimation methods developed so far, stem from a deterministic framework. Capitalizing on the celebrated least absolute shrinkage and selection operator (lasso) \cite{1996_tibshirani}, an $\ell_1$ regularization term is introduced in the cost function of these methods. In this context, by incorporating an $\ell_1$ (or a log-sum) penalty term in the cost function of the standard least mean square (LMS) algorithm, adaptive LMS algorithms  that are able to recursively identify sparse systems are derived in \cite{2009_chen}. Inclusion of an $\ell_1$ regularization factor or a more general regularizing term in the least squares (LS) cost function has also been proposed in \cite{2010_angelosante} and \cite{2011_eksioglu}, respectively. In \cite{2010_angelosante} adaptive coordinate-descent type algorithms are developed with sparsity being imposed via soft-thresholding, while in \cite{2011_eksioglu} recursive LS (RLS) type schemes are designed. 
An $\ell_1$ regularized RLS type algorithm that utilizes the expectation maximization (EM) algorithm as a low-complexity solver is described in \cite{2010_babadi}. In a different spirit, a subgradient projection-based adaptive algorithm that induces sparsity using projections on weighted $\ell_1$ balls is developed and analyzed in \cite{2011_kopsinis}. Adaptive greedy variable selection schemes have been also recently reported, e.g. \cite{2010_mileounis}. However, these algorithms require, at least, a rough knowledge of the signal sparsity level and work effectively for sufficiently high signal sparsity.

In this paper, we depart from the deterministic setting adopted so far in previous works and deal with the sparse adaptive estimation problem within a Bayesian framework. In such a framework, a Bayesian model is first defined comprising, a) a likelihood function specified by the assumed measurement data generation process and b) prior distributions for all model parameters, (which are thus considered as random variables), properly chosen to adhere to the constraints of the problem. In particular, to induce sparsity, suitable heavy-tailed sparsity promoting priors are assigned to the weight parameters of interest. 
Then a variational Bayesian inference method is utilized to approximate the joint posterior distribution of all model parameters, from which estimates of the sought parameters can be obtained via suitably defined iterative algorithms. It should be emphasized though that the various Bayesian inference methods are designed to solve the \emph{batch} estimation problem, i.e. they provide the parameter estimates based on a given fixed size block of data and observations.

In the context described above, the contribution of this work is twofold. First, we provide a unified derivation of a family of Bayesian batch estimation techniques. Such a derivation passes through a) the selection of a generalized prior distribution for the \emph{sparsity inducing parameters} of the model and b) the adoption of a mean-field variational approach \cite{1999_jordan,2000_jaakkola,2006_bishop} to perform Bayesian inference. The adopted fully factorized variational approximation method relies on an independence assumption on the posterior of all involved model parameters and leads to simple sparsity aware iterative batch estimation schemes with proven convergence. The derivation of the above batch estimation algorithms constitutes the prerequisite step that facilitates the deduction of the novel adaptive variational Bayes algorithms, which marks the second contribution and main objective of this work. 
The proposed adaptive algorithms consist of two parts, namely, a common part encompassing  time 
update formulas of the basic model parameters and a sparsity enforcing mechanism, which is different for the various Bayesian model priors assumed. The algorithms are numerically robust and are based on second order statistics having a computational complexity similar to that of other related sparsity aware deterministic schemes. Moreover, extensive simulations under various time-varying conditions show that they converge faster to sparse solutions and offer, in principle, lower steady-state estimation error compared to existing algorithms. The major advantage, though, of the proposed algorithms is that thanks to their Bayesian origin, they are fully automated. While related sparse deterministic algorithms (in order to achieve optimum performance) involve application- and conditions-dependent regularization parameters that need to be predetermined via exhaustive fine-tuning, the Bayesian algorithms presented in this paper directly infer all model parameters from the data, and hence, the need for parameter 
fine-tuning 
is entirely eliminated. 
This, combined with their robust sparsity inducing properties, makes them particularly attractive for use in practice. A preliminary version of a part of this work has been presented in \cite{2013_themelis}\footnote{Note that a Bayesian approach to adaptive {\it filtering} has been previously proposed in \cite{2003_koeppl}. However, in \cite{2003_koeppl} a type-II maximum likelihood inference method is adopted that leads to a regularized RLS-type scheme. This is completely different from the approach and algorithms described in this work.}.  

The rest of the paper is organized as follows. Section \ref{sec:prob_form} defines the mathematical formulation of the adaptive estimation problem from a LS point of view. In Section \ref{sec:Bays_model} the adopted hierarchical Bayesian model is described. A family of batch variational Bayes iterative schemes is presented in Section \ref{sec:bayes_inference}. The new sparse adaptive variational Bayes algorithms are developed in Section \ref{sec:sbl_ad_filtering}. In Section \ref{sec:analysis} an analysis of the proposed algorithms is presented and their relation to other known algorithms is established. Extensive experimental results are provided in Section \ref{sec:experiments} and concluding remarks are given in Section \ref{sec:conclusion}.

\emph{Notation}: Column vectors are represented as boldface lowercase letters, e.g. ${\mathbf x}$, and matrices as boldface uppercase letters, e.g. ${\mathbf X}$, while the $i$-th component of vector $\mathbf x$ is denoted by $x_i$ and the $ij$-th element of matrix $\mathbf X$ by $x_{ij}$. Moreover, $(\cdot)^T$ denotes transposition, $\|\!\cdot\!\|_1$ stands for the $\ell_1$-norm, $\|\!\cdot\!\|$ stands for the standard $\ell_2$-norm, $|\!\cdot\!|$ denotes the determinant of a matrix or absolute value in case of a scalar, $\mathcal N(\cdot)$ is the Gaussian distribution, $\mathcal {G}(\cdot)$ is the Gamma distribution, $\mathcal {IG}(\cdot)$ is the inverse Gamma distribution, $\mathcal {GIG}(\cdot)$ is the generalized inverse Gaussian distribution, $\Gamma(\cdot)$ is the Gamma function, 
$\langle \cdot \rangle$ is the expectation operator, $\rm{diag}({\mathbf x})$ denotes a diagonal matrix whose diagonal entries are the elements of ${\mathbf x}$, and $\rm{diag}({\mathbf X})$ is a column vector containing the main diagonal elements of a square matrix $\mathbf{X}$. 
Finally, we use the semicolon $(;)$ and the vertical bar $(|)$ characters to express the dependence of a random variable on parameters and other random variables, respectively.

\section{Problem statement}

\label{sec:prob_form}

Let $\mathbf{w}(n) = [w_1(n), w_2(n), \ldots, w_N(n)]^T \in \mathbb{R}^N$ denote a \emph{sparse} time-varying weight vector having $\xi \ll N$ non-zero elements, where $n$ is the time index. We wish to estimate and track $\mathbf{w}(n)$ in time by observing a stream of sequential data obeying to the following linear regression model, 
\begin{align}
y(n) = \mathbf{x}^T(n) \mathbf{w}(n) + \epsilon(n),
\label{eq:model}
\end{align}
where ${\mathbf x}(n)=[x_1(n),x_2(n),\ldots,x_N(n)]^T$ is a \emph{known} $N \times 1$ regression vector, and $\epsilon(n)$ denotes the uncorrelated with ${\mathbf x}(n)$ added Gaussian noise of zero mean and variance $\beta^{-1}$ (or precision $\beta$), i.e. $\epsilon(n) \sim {\cal N}(\epsilon(n)|0,\beta^{-1})$. The linear data generation model given in (\ref{eq:model}) fits very well or, at least, approximates adequately the hidden mechanisms in many signal processing tasks. 
Let 
\begin{align}
{\mathbf y}(n) = [y(1), y(2), \dots, y(n)]^T
 \label{eq:yn}
\end{align}
and  
\begin{align} 
\mathbf{X}(n) = [\mathbf{x}(1), \mathbf{x}(2), \dots, \mathbf{x}(n)]^T \label{eq:Xn} 
\end{align} 
be the $n \times 1$ vector of observations and the $n \times N$ input data matrix respectively, up to time $n$. Then, the unknown weight vector ${\mathbf w}(n)$ can be estimated by minimizing with respect to (w.r.t.) $\hat {\mathbf w}(n)$ the following exponentially weighted LS cost function\footnote{Note that a fixed size sliding in time data window could be also used.},  
\begin{align} 
\mathcal{J}_{\mathrm{LS}}(n) & =  \sum_{j=1}^n \lambda^{n-j} |y(j) - {\mathbf x}^T(j)  \hat{{\mathbf w}}(n)|^2  =  \| \boldsymbol \Lambda^{1/2}(n) \mathbf{y}(n) - \boldsymbol \Lambda^{1/2}(n) {\mathbf X}(n)  \hat{{\mathbf w}}(n) \|^2. 
\label{eq:rls_cost_function} 
\end{align} 
The parameter $\lambda$, $0 \ll \lambda \leq 1$, is commonly referred to as the \emph{forgetting factor} (because it weights more heavily recent data and `forgets' gradually old data), and $\boldsymbol \Lambda(n) = \mathrm{diag}([\lambda^{n-1}, \lambda^{n-2}, \dots, 1 ]^T)$. It is well-known that the vector $\hat {\mathbf w}(n)$ that minimizes $\mathcal{J}_{\mathrm{LS}}(n)$ is given by the solution of the celebrated normal equations \cite{2001_haykin}. In an adaptive estimation setting, the cost function in (\ref{eq:rls_cost_function}) can be optimized recursively in time by utilizing the RLS algorithm. The RLS algorithm, a) reduces the computational complexity from $\mathcal{O}(N^3)$, which is required for solving the normal equations per time iteration, to $\mathcal{O}(N^2)$, b) has constant memory requirements despite the fact that the size of the data grows with $n$, and, c) has the ability of tracking possible variations 
of $\mathbf{w}(n)$ as $n$ 
increases.


However, the RLS algorithm does not specifically exploit the inherent sparsity of the parameter vector $\mathbf{w}(n)$, so as to improve its initial convergence rate and estimation performance. To deal with this issue, a number of adaptive deterministic LS-type algorithms have been recently proposed, e.g. \cite{2010_angelosante,2011_eksioglu,2010_babadi}. In all these schemes, the LS cost function is supplemented with a regularization term that penalizes the $\ell_1$-norm of the unknown weight vector, i.e., 
\begin{align} 
\mathcal{J}_{\mathrm{LS-}\ell_1}(n) = \| \boldsymbol \Lambda^{1/2}(n) \mathbf{y}(n) - \boldsymbol \Lambda^{1/2}(n) {\mathbf X}(n)  \hat{{\mathbf w}}(n) \|^2 + \tau \| \hat{{\mathbf w}}(n)\|_1, 
\label{eq:rls_l1__cost_function} 
\end{align} 
where $\tau > 0$ is a regularization parameter controlling the sparsity of $\hat {\mathbf w}(n)$, that should be properly selected. Regularization with the  $\ell_1$-norm has its origin in the widely known lasso operator, \cite{1996_tibshirani}, and is known to promote sparse solutions. 

In this paper, unlike previous studies, we provide an analysis of the sparse adaptive estimation problem from a Bayesian perspective. In this framework, we derive a class of variational Bayes estimators that are built upon hierarchical Bayesian models featuring heavy-tailed priors. A basic characteristic of heavy-tailed priors is their sparsity inducing nature. These prior distributions are known to improve robustness of regression and classification tasks to outliers and have been widely used in variable selection problems, \cite{2011_bioucas-dias,2001_girolami}. The variational Bayesian inference approach adopted in this paper, a) exhibits low computational complexity compared to (the possible alternative) Markov Chain Monte Carlo (MCMC) sampling method, \cite{2006_bishop}, and b) performs inference for all model parameters, including the sparsity promoting parameter $\tau$, as opposed to deterministic methods. In the following, we analyze a general hierarchical Bayesian model for the batch estimation 
problem first (i.e., when $n$ is considered fixed), and then we show how the proposed variational Bayes inference method can be extended in an adaptive estimation setting. 

\section{Bayesian modeling} 
\label{sec:Bays_model}  

To simplify the description of the hierarchical Bayesian model we temporarily drop the dependence of all model quantities from the time indicator $n$.  Time dependency will be re-introduced in Section \ref{sec:sbl_ad_filtering}, where the proposed adaptive variational schemes are presented. To consider the estimation problem at hand from a Bayesian point of view, we first define a likelihood function based on the given data generation model and then we introduce sparsity to our estimate by assigning a suitable heavy-tailed prior distribution over the parameter vector ${\mathbf w}$. In order to account for the exponentially weighted data windowing used in (\ref{eq:rls_cost_function}), the following observation model is considered 
\begin{align} 
{\mathbf y} = {\mathbf X}{\mathbf w} +\boldsymbol \varepsilon 
\label{eq:altmodel} 
\end{align} 
where $\boldsymbol \varepsilon\sim {\cal N}(\boldsymbol \varepsilon|{\mathbf 0},\beta^{-1}\boldsymbol \Lambda^{-1})$. From this observation model and the statistics of the noise vector $\boldsymbol \varepsilon$, it turns out that the corresponding likelihood function is 
\begin{align} 
p({\mathbf y}|{\mathbf w},\beta) = (2 \pi)^{-\frac{n}{2}} \beta^{\frac{n}{2}}  |\boldsymbol \Lambda|^{\frac{1}{2}} {\rm exp}\left[ - \frac{\beta}{2} \| \boldsymbol \Lambda ^{\frac{1}{2}}{\mathbf y} - \boldsymbol \Lambda ^{\frac{1}{2}} \mathbf{X} {\mathbf w}\|^2\right]. 
\label{eq:likelihood} 
\end{align} 
Notice that the maximum likelihood estimator of (\ref{eq:likelihood}) coincides with the LS estimator that minimizes (\ref{eq:rls_cost_function}). However, as mentioned previously, our estimator should be further constrained to be sparse. To this end, the likelihood is complemented by suitable {\it conjugate} priors w.r.t. (\ref{eq:likelihood}) over the parameters $\mathbf{w}$ and $\beta$ \cite{2003_figueiredo,2008_park}. The prior for the noise precision $\beta$ is selected to be a Gamma distribution with parameters $\rho$ and $\delta$, i.e., 
\begin{align} 
p(\beta;\rho,\delta) = {\cal G}(\beta;\rho,\delta) = \frac{\delta^\rho}{\Gamma(\rho)}\beta^{\rho-1} \mathrm{exp} \left[- \delta \beta \right].
\label{eq:betaprior} 
\end{align} 
Next, a two-level hierarchical heavy-tailed prior is selected for the parameter vector $\mathbf{w}$, that reflects our knowledge that many of its components are zero or nearly zero. In the first level of hierarchy, a Gaussian prior is attached on $\mathbf{w}$, i.e.,  
\begin{align} 
p({\mathbf w}|\boldsymbol \alpha,\beta) & = \mathcal{N}(\mathbf{w}|0,\beta^{-1} \boldsymbol {\mathbf A}^{-1}) = \prod_{i=1}^{N} p(w_i|\alpha_i,\beta)  = \prod_{i=1}^{N} \left(2\pi\right)^{-\frac{1}{2}} \beta ^{\frac{1}{2}} \alpha_i^\frac{1}{2} \mathrm{exp}\left[ - \frac{\beta}{2} w_i^2 \alpha_i \right]. 
\label{eq:wprior} 
\end{align} 
where $\boldsymbol \alpha = [\alpha_1,\alpha_2,\ldots,\alpha_N]^T$ is the vector of the precision parameters of the $w_i$'s,  ${\mathbf A}=\mathrm{diag}(\boldsymbol \alpha)$ and the $w_i$'s have been assumed a priori independent. Now, depending on the choice of the prior distribution for the precision parameters in $\boldsymbol \alpha$ at the second level of hierarchy, various heavy-tailed distributions may arise for ${\mathbf w}$, such as the Student-t or the Laplace distribution. To provide a unification of all these distributions in a single model, we assume that the sparsity enforcing parameters $\alpha_i$ follow a generalized inverse Gaussian (GIG) distribution, expressed as\footnote{More general models can be found in \cite{2012_zhang}.} 
\begin{align}  
p(\alpha_i ; a,b,c) & = \mathcal{G}\mathcal{I}\mathcal{G}(\alpha_i ; a,b,c)  =\frac{(a \slash b)^{(c\slash 2 )}}{2 \mathcal{K}_c(\sqrt{ab})} \alpha_i^{c-1} \mathrm{exp}[-\frac{1}{2}(a \alpha_i + b \slash \alpha_i)] ,
\label{eq:gig}
\end{align} 
where $a,b \geq 0, c \in \mathbb{R}$ and $\mathcal{K}_{c}(\cdot) $ is the modified Bessel function of the second kind. Our analysis will be developed for a special case of (\ref{eq:gig}), which arises by setting $b=0, a > 0,$  and $c>0$ \footnote{Note that for $b=0, a > 0, c> 0$ the GIG distribution is well defined, \cite{2012_zhang}.}. Other special cases of (\ref{eq:gig}) that also lead to sparse estimates for ${\mathbf w}$ will be discussed in the next Section. Hence, by selecting $b=0$ and $c>0$, it can be shown that (\ref{eq:gig}) becomes a Gamma distribution with scale parameter $c$ and rate parameter $a/2$, 
\begin{align}  
p(\alpha_i ; c,a) = \mathcal{G} (\alpha_i ; c,a\slash 2) = \frac{(a/2)^c}{\Gamma(c)}\alpha_i^{c-1} {\rm exp} \left[- \frac{a}{2} \alpha_i \right],
\label{eq:alpha_i_prior} 
\end{align}
for $i=1,2,\ldots,N$. If we integrate out the precision parameter $\boldsymbol \alpha$ from (\ref{eq:wprior}) using (\ref{eq:alpha_i_prior}) as a prior, it is easily verified that the two-level hierarchical prior defined by (\ref{eq:wprior}) and (\ref{eq:alpha_i_prior}) is equivalent to assigning a Student-t distribution over the parameter vector $\mathbf{w}$, which depends on the hyperparameters $c$ and $a$, \cite{2001_tipping}.  Besides, the prior in (\ref{eq:alpha_i_prior}) for each $\alpha_i, i=1,2,\ldots,N$, is the conjugate pair of $q(w_i|\alpha_i,\beta)$ in (\ref{eq:wprior}). The existence of conjugacy among the distributions of the Bayesian hierarchical model is crucial for the development of a computationally efficient variational inference approach.  

It should be noted here that not all members of the GIG class of distributions lead to conjugacy among the distributions of the Bayesian hierarchical model. The Bayesian model presented above (Eqs. (\ref{eq:likelihood}), (\ref{eq:betaprior}), (\ref{eq:wprior}), (\ref{eq:alpha_i_prior})) is similar to that proposed in \cite{2001_tipping}, except for the normalization by $\beta$ of the variances of $w_i$'s in (\ref{eq:wprior}). However, it can be shown that the modification adopted here ensures the unimodality of the posterior joint distribution, \cite{2008_park}, and leads to simpler and more compact parameter update expressions, as will be seen later.     

\section{Mean-Field Variational Bayesian inference} 
\label{sec:bayes_inference} 

So far we have presented a generative model  for the observations data (Eq. (\ref{eq:altmodel})) and a hierarchical Bayesian model (Eqs. (\ref{eq:betaprior}), (\ref{eq:wprior}), (\ref{eq:alpha_i_prior})), treating the model parameters as random variables. To proceed with Bayesian inference, the computation of the joint posterior distribution over the model parameters is required. Using Bayes' law, this distribution is expressed as 
\begin{align} 
p(\mathbf{w},\beta, \boldsymbol \alpha | \mathbf{y}) = \frac{p(\mathbf{y},\mathbf{w},\beta, \boldsymbol \alpha) }{\int p(\mathbf{y},\mathbf{w},\beta, \boldsymbol \alpha) d\mathbf{w} d\beta d\boldsymbol \alpha }.  
\label{eq:posterior_exact} 
\end{align} 
However, due to the complexity of the model, we cannot directly compute the posterior of interest, since the integral in (\ref{eq:posterior_exact}) can not be expressed in closed form. Thus, we resort to approximations. In this paper, we adopt the variational framework, \cite{1999_jordan,2000_jaakkola,2000_attias,2008_tzikas,2003_beal}, to approximate the posterior in (\ref{eq:posterior_exact})  with a simpler, \emph{variational} distribution  $q(\mathbf{w},\beta, \boldsymbol \alpha)$. From an optimization point of view, the parameters of $q(\mathbf{w},\beta, \boldsymbol \alpha)$ are selected so as to minimize the Kullback-Leibler divergence metric between the true posterior $p(\mathbf{w},\beta, \boldsymbol \alpha |\mathbf{y})$ and the variational distribution  $q(\mathbf{w},\beta, \boldsymbol \alpha)$ \cite{2008_tzikas}. This minimization is equivalent to maximizing the \emph{evidence lower bound} (ELBO) (which is a lower bound on the logarithm of the data marginal likelihood $\mathrm{log}p(\mathbf{y})$) w.
r.t. the 
variational distribution $q(\mathbf{w},\beta,\boldsymbol \alpha)$, \cite{2003_beal}.  
Based on the mean-field theory from statistical physics \cite{1987_peterson}, we constrain $q(\mathbf{w},\beta, \boldsymbol \alpha)$ to the family of distributions, which are {\it fully} factorized w.r.t. their parameters yielding  
\begin{align} 
q(\mathbf{w},\beta, \boldsymbol \alpha) & = q(\mathbf{w}) q(\beta) q(\boldsymbol \alpha)  = \prod_{i=1}^{N} q(w_i) q(\beta) \prod_{i=1}^{N} q(\alpha_i),
\label{eq:qfactors}
\end{align} 
that is, all model parameters are assumed to be \emph{a posteriori} independent. This fully factorized form of the approximating distribution $q(\mathbf{w},\beta,\boldsymbol \alpha)$ turns out to be very convenient, mainly because it results to an optimization problem that is computationally tractable. 
In fact, if we let $\theta_i$ denote the $i$-th component of the vector $\boldsymbol \theta = [w_1,\ldots,w_N,\beta,\alpha_1,\ldots,\alpha_N]^T$ containing the parameters of the Bayesian hierarchical model, maximization of the ELBO results in the following expression for $q(\theta_i)$, \cite{2006_bishop}, 
\begin{align} 
q(\theta_i) = \frac{\mathrm{exp} \left[ \langle \mathrm{log} p(\mathbf{y},\boldsymbol \theta) \rangle_{j \neq i}  \right] }{\int \mathrm{exp} \left[ \langle \mathrm{log} p(\mathbf{y},\boldsymbol \theta) \rangle_{j \neq i}  \right] d\theta_i},
\label{eq:gen_qtheta} 
\end{align}
where $\langle \cdot \rangle_{j \neq i}$ denotes the expectation w.r.t. $\prod_{j \neq i} q(\theta_j)$. Note that this is not a closed form solution, since every factor $q(\theta_i)$ depends on the remaining  factors $q(\theta_j)$, for $j \neq i$. 
However, the interdependence between the factors $q(\theta_i)$ gives rise to a cyclic optimization scheme, where the factors are initialized appropriately, and each one is then updated via (\ref{eq:gen_qtheta}), by holding the remaining factors fixed. Each update cycle is known to increase the ELBO until convergence, \cite{2003_beal}.  

Applying (\ref{eq:gen_qtheta}) to the proposed model (exact computations are reported in Appendix \ref{app:q_w_i}), the approximating distribution for each coordinate $w_i, i=1,2,\ldots,N$, is found to be Gaussian,  
\begin{align} 
q(w_i) = {\cal N}(w_i;\mu_i,\sigma_i^2)=(2 \pi)^{-\frac{1}{2}} \sigma_i^{-1} \mathrm{exp} \left[ -\frac{1}{2} \frac{(w_i-\mu_i)^2}{\sigma_i^2}  \right], 
\label{eq:qwi} 
\end{align} 
with parameters $\mu_i$ and $\sigma_i^2$ given by 
\begin{align} 
\sigma_i^2 &= \langle \beta \rangle ^{-1} ( \mathbf{x}_i^T \boldsymbol \Lambda \mathbf{x}_i + \langle \alpha_i \rangle )^{-1}, \label{eq:sigmai} \\ 
\mu_i &= \langle  \beta \rangle \sigma_i^2 \mathbf{x}_i^T \boldsymbol \Lambda  ( {\mathbf y} - \mathbf{X}_{\neg i} \boldsymbol \mu_{\neg i }).  
\label{eq:mui} 
\end{align} 
In (\ref{eq:mui}),  $\mathbf{X}_{\neg i}$ results from the data matrix $\mathbf{X}$ after removing its $i$-th column $\mathbf{x}_i$, $\boldsymbol \mu = [\mu_1, \mu_2, \dots, \mu_N]^T$ is the posterior mean value of $\mathbf{w}$,  $\boldsymbol \mu_{\neg i }$ results from $\boldsymbol \mu$ after the exclusion of its $i$-th element, and expectation $\langle \cdot \rangle$ is w.r.t. the variational  distributions $q(\cdot)$ of the parameters appearing within each pair of brackets.  Notice that each element $w_i$ of $\mathbf{w}$ is treated separately and thus $q(w_i)$ constitutes an individual factor in the rightmost hand side of (\ref{eq:qfactors}), as opposed to having a single factor $q(\mathbf{w})$ for the whole vector $\mathbf{w}$, as in \cite{2011_shutin}; this is beneficial for the development of the adaptive schemes that will be presented in the next Section.   Working in a similar manner for the noise precision $\beta$, we get that $q(\beta)$ is a Gamma distribution expressed as  
\begin{align} 
q(\beta)  = \mathcal{G}(\beta;\tilde{\rho},\tilde{\delta}) = \frac{\tilde{\delta}^{\tilde{\rho}}}{\Gamma (\tilde{\rho})} \beta ^{ \tilde{\rho} -1 } \mathrm{exp} \left[- \tilde{\delta} \beta \right],
\label{eq:qbeta} 
\end{align} 
with $\tilde{\rho} = \frac{n+N}{2} + \rho$ and $\tilde{\delta} = \delta+\frac{1}{2} \left \langle \| \boldsymbol \Lambda^{\frac{1}{2}} {\mathbf y} - \boldsymbol \Lambda^{\frac{1}{2}} \mathbf{X} {\mathbf w}\|^2\right \rangle + \frac{1}{2} \left  \langle  {\mathbf w} ^T {\mathbf A}{\mathbf w} \right \rangle$.  

Finally, the variational distribution of the precision parameters $\alpha_i$'s turns out to be also Gamma given by,  
\begin{align} 
q(\alpha_i) = \frac{\left( \frac{\langle \beta \rangle \langle w_i^2 \rangle}{2} + a \right)^{c+\frac{1}{2}}}{\Gamma(c+\frac{1}{2})} \alpha_i^{c-\frac{1}{2}} \mathrm{exp} \left[ - \left( \frac{\langle \beta \rangle \langle w_i^2 \rangle  }{2} + a \right) \alpha_i \right].  
\label{eq:qalphai} 
\end{align} 
Owing to the conjugacy of our hierarchical Bayesian model, the variational distributions in (\ref{eq:qwi}), (\ref{eq:qbeta}), and (\ref{eq:qalphai}) are expressed in a standard exponential form. Notice also that the  parameters of all variational distributions are expressed in terms of expectations of expressions of the other parameters. More specifically from (\ref{eq:qbeta}) and (\ref{eq:qalphai}) we get   %
\begin{align} 
& \langle \beta \rangle = \frac{ \frac{n+N}{2} + \rho }{  \delta + \frac{1}{2} \left \langle \|\boldsymbol \Lambda^{\frac{1}{2}}{\mathbf y} - \boldsymbol \Lambda^{\frac{1}{2}}\mathbf{X} \mathbf{w}\|^2 \right \rangle  + \frac{1}{2} \sum_{i=1}^N \langle w_i^2 \rangle  \langle \alpha_i \rangle  }  \label{eq:beta_exp}, \end{align} \begin{align} \langle \alpha_i \rangle = \frac{2c+1}{a +  \langle \beta \rangle \langle w_i^2 \rangle} 
\label{eq:alpha_i_exp}, 
\end{align} 
for $i=1,2,\ldots,N$ respectively, whereas $\langle w_i \rangle \equiv \mu_i, i=1,2,\ldots,N$ is given by (\ref{eq:mui}). In addition, since $\langle w_i^2 \rangle =  \mu_i^2 + \sigma_i^2$, it can be easily shown that the middle term in the denominator of the right hand side (RHS) of (\ref{eq:beta_exp}) is evaluated as  
\begin{align} 
\left \langle \| \boldsymbol \Lambda^{\frac{1}{2}} {\mathbf y} -\boldsymbol \Lambda^{\frac{1}{2}} \mathbf{X} \mathbf{w}\|^2 \right \rangle  =  \| \boldsymbol \Lambda^{\frac{1}{2}} {\mathbf y} - \boldsymbol \Lambda^{\frac{1}{2}} \mathbf{X} \boldsymbol \mu\|^2  + \sum_{i=1}^N \sigma_i^2 \mathbf{x}_i^T \boldsymbol \Lambda \mathbf{x}_i. 
\label{eq:beta_exp_den} 
\end{align} 
The final variational scheme involves updating (\ref{eq:sigmai}), (\ref{eq:mui}), for $i=1,2,\ldots,N$, (\ref{eq:beta_exp}), and (\ref{eq:alpha_i_exp}) for $i=1,2,\ldots,N$ in a sequential manner. The hyperparameters $\rho,\delta,c,a$ are set to very small values corresponding to almost noninformative priors for $\beta$ and the $\alpha_i$'s. The variational algorithm solves the batch estimation problem defined in (\ref{eq:rls_cost_function}) and due to the convexity of the factors $q(w_i),q(\beta)$ and $q(\alpha_i)$ converges to a sparse solution in a few cycles \cite{2006_bishop}. The final estimate of the sparse vector $\mathbf{w}$ is the mean $\boldsymbol \mu$ of the approximating posterior $q(\mathbf w)$.  A summary of the sparse variational Bayes procedure utilizing a hierarchical Student-t prior is shown in Table \ref{tab:VEM_student}. 
The Table includes a  description of the initial Bayesian model, the resulting variational distributions and the corresponding sparse variational Bayes Student-t based (SVB-S) iterative scheme.  

\subsection{Batch variational Laplace algorithms} 
\label{subs:batch_l} 

As mentioned in Section \ref{sec:Bays_model}, depending on the choice of the prior for the precision parameters $\alpha_i$'s, various sparsity inducing prior distributions may arise for ${\mathbf w}$. Such a prior is obtained by setting $a=0$ and $c=-1$ in (\ref{eq:gig}), in which case the following {\it inverse} Gamma distribution is obtained 
\begin{align} 
p(\alpha_i|b) = \mathcal{IG}(\alpha_i|1,\frac{b}{2}) = \frac{b}{2}  \alpha_i^{-2}{\rm exp} \left[- \frac{b}{2} \frac{1}{\alpha_i} \right]  
\label{eq:ig} 
\end{align}  
for $i=1,2,\ldots,N$, while $b$ is assumed to follow a conjugate Gamma distribution, $b \sim {\cal G}(b;\kappa,\nu)$. As shown in Appendix \ref{app:laplpr}, if we integrate out $\boldsymbol \alpha$ from the hierarchical prior of ${\mathbf w}$ defined by (\ref{eq:wprior}) and (\ref{eq:ig}), a sparsity promoting multivariate Laplace distribution arises for ${\mathbf w}$. In addition, it can be shown that the resulting Bayesian model preserves an equivalence relation with the lasso \cite{1996_tibshirani} in that its maximum a posteriori probability (MAP) estimator coincides with the vector that minimizes the lasso criterion \cite{2003_figueiredo, 2010_babacan}\footnote{Note, however, that in \cite{2003_figueiredo,2010_babacan} a different, (in terms of the parametres that impose sparsity), model is described. Specifically, instead of the precisions $\alpha_i$'s of $w_i$'s, their variances $\gamma_i$'s are used, with $\gamma_i = \alpha_i^{-1}$, on which Gamma priors of the form ${\cal G}(\gamma_i|1,\frac{b}{2})$ 
are 
assigned.}. A summary of this alternative model including a description of the resulting sparse variational Bayes iterative scheme based on a Laplace prior (SVB-L), is shown in Table \ref{tab:VEM_Laplace}. Note from Table \ref{tab:VEM_Laplace} that the variational distribution $q(\alpha_i)$ now becomes the following GIG distribution
\begin{align}
q(\alpha_i) = {\cal G}{\cal I}{\cal G}\left(\alpha_i;\langle \beta \rangle \langle w_i^2 \rangle,\langle b \rangle,-1/2\right).
\label{eq:qa_i}
\end{align}
Moreover, except for the  mean of $\alpha_i$
\begin{align} 
\langle \alpha_i \rangle = \sqrt{\frac{\langle b \rangle}{\langle \beta \rangle \langle w_i^2 \rangle}},  
\label{eq:ailap} 
\end{align} 
the expectation $\langle \frac{1}{\alpha_i} \rangle$ w.r.t. $q(\alpha_i)$ is also required, which can be expressed as  
\begin{align} 
\left \langle \frac{1}{\alpha_i} \right \rangle = \langle \gamma_i  \rangle = \frac{1}{\langle \alpha_i \rangle} + \frac{1}{\langle b \rangle}.  
\label{eq:invailap} 
\end{align}  
As noted in \cite{2010_griffin}, the single shrinkage parameter $b$ penalizes both zero and non-zero coefficients equally and it is not flexible enough to express the variability of sparsity among the unknown weight coefficients. In many circumstances, this leads to limited posterior inference and, evidently, to poor estimation performance. Hence, a slightly  different (but, as it will be shown, much more powerful) model can be constructed by allowing for multiple parameters $b_i$, one for each $\alpha_i$ in (\ref{eq:ig}), i.e,  
\begin{align} 
p(\alpha_i|b_i) = \mathcal{IG}(\alpha_i|1,\frac{b_i}{2}) = \frac{b_i}{2}  \alpha_i^{-2}{\rm exp} \left[- \frac{b_i}{2} \frac{1}{\alpha_i} \right] 
\label{eq:igbi} 
\end{align} 
and $b_i \sim {\cal G}(b_i;\kappa,\nu)$, \cite{2012_themelis}. Working as in Appendix \ref{app:laplpr}, it is easily shown that for such a prior for $\alpha_i$'s, the resulting prior for ${\mathbf w}$ is a multivariate, {\it multi-parameter} Laplace distribution (each $b_i$ corresponds to a single $w_i$). Furthermore, the MAP estimator for this model is identical to the vector that minimizes the so-called adaptive (or weighted) lasso cost function \cite{2006_zou, 2012_themelis,2012_themelis1}. A summary of the proposed sparse variational Bayes scheme, which is based on a multiparameter Laplace prior  (SVB-mpL) is shown in Table \ref{tab:VEM_multiLaplace}.   

By inspecting Tables \ref{tab:VEM_student}, \ref{tab:VEM_Laplace} and \ref{tab:VEM_multiLaplace}, we see that SVB-S, SVB-L and SVB-mpL share common rules concerning the computation of the ``high in the hierarchy''  model parameters ${\mathbf w}, \beta$, while they differ in the way the sparsity imposing precision parameters in $\boldsymbol \alpha$ are computed. To the best of our knowledge, it is the first time that these three schemes are derived via a mean-field fully factorized variational Bayes inference approach, under a unified framework. Such a presentation not only highlights their common features and differences, but it also facilitates a unified derivation of the corresponding adaptive algorithms that will be described in the next Section.   
\begin{table*}[t] 
\centering 
\newcolumntype{R}{>{\raggedleft\arraybackslash}X}%
\begin{tabularx}{1.2 \textwidth}{ |l|R| }   
\hline   
\multicolumn{2}{|l|}{\textbf{Sparse variational Bayes with a Student-t prior}} \\  
\hline  
Data likelihood & $p({\mathbf y}|{\mathbf w},\beta) = (2 \pi)^{-\frac{n}{2}} \beta^{\frac{n}{2}}  |\boldsymbol \Lambda|^{1/2} {\rm exp}\left[ - \frac{\beta}{2} \| \boldsymbol \Lambda ^{1/2}{\mathbf y} - \boldsymbol \Lambda ^{1/2} \mathbf{X} {\mathbf w}\|^2\right]$  \\   
\hline   
Prior distributions  & $p(\beta;\rho,\delta) = \mathcal{G}(\beta;\rho,\delta) = \frac{\delta^\rho}{\Gamma(\rho)}\beta^{\rho-1} \mathrm{exp} \left[- \delta \beta \right]$ \\   ~ & $p({\mathbf w}|\boldsymbol \alpha,\beta) = \mathcal{N}(\mathbf{w}|0,\beta^{-1} \mathbf{A}^{-1}) = \prod_{i=1}^{N} \left(2\pi\right)^{-\frac{1}{2}} \beta ^{\frac{1}{2}} \alpha_i^\frac{1}{2} \mathrm{exp}\left[ - \frac{\beta}{2} w_i^2 \alpha_i \right]$  \\ 
~ & $p(\alpha_i;c,a) = \mathcal{G}(\alpha_i;c,a/2) = \frac{(a/2)^c}{\Gamma(c)}\alpha_i^{c-1} {\rm exp} \left[- \frac{a}{2} \alpha_i \right], i = 1, 2, \dots, N$ \\  
\hline  
Variational distributions  & $q(\beta) =\mathcal{G}(\beta;\tilde{\rho},\tilde{\delta})$, with $\tilde{\rho} = \frac{n+N}{2} + \rho$ and $\tilde{\delta} = \delta +\frac{1}{2} \left \langle \| \boldsymbol \Lambda^{\frac{1}{2}} {\mathbf y} - \boldsymbol \Lambda^{\frac{1}{2}} \mathbf{X} {\mathbf w}\|^2\right \rangle + \frac{1}{2} \left  \langle  {\mathbf w} ^T {\mathbf A}{\mathbf w} \right \rangle$ \\
~ & $q(w_i) = \mathcal{N}(w_i;\mu_i,\sigma_i^2)$, with $\sigma_i^2 = \langle \beta \rangle ^{-1} ( \mathbf{x}_i^T \boldsymbol \Lambda \mathbf{x}_i + \langle \alpha_i \rangle )^{-1}$ and $\mu_i = \langle  \beta \rangle \sigma_i^2 \mathbf{x}_i^T \boldsymbol \Lambda  ( {\mathbf y} - \mathbf{X}_{\neg i} \boldsymbol \mu_{\neg i }), i = 1, 2, \dots, N$\\ 
~  & $q(\alpha_i) = \mathcal{G}(\beta;\tilde{c},\tilde{a})$, with $\tilde{c} = c+\frac{1}{2}$ and $\tilde{a} = \frac{\langle \beta \rangle \langle w_i^2 \rangle + a}{2}, i = 1, 2, \dots, N $ \\ 
\hline   
Variational updates &  $\langle \beta \rangle = ( n+N  + 2 \rho ) \slash ( 2 \delta +  \left \langle \|\boldsymbol \Lambda^{1/2}{\mathbf y} - \boldsymbol \Lambda^{1/2}\mathbf{X} \mathbf{w}\|^2 \right \rangle  +  \sum_{i=1}^N \left \langle w_i^2 \right \rangle  \langle \alpha_i \rangle  ) $ \\   ~ & $\sigma_i^2 = \langle \beta \rangle ^{-1} ( \mathbf{x}_i^T \boldsymbol \Lambda \mathbf{x}_i + \langle \alpha_i \rangle )^{-1}, i = 1, 2, \dots, N$ \\   ~ & $\langle w_i \rangle \equiv \mu_i = \langle  \beta \rangle \sigma_i^2 \mathbf{x}_i^T \boldsymbol \Lambda  ( {\mathbf y} - \mathbf{X}_{\neg i} \boldsymbol \mu_{\neg i }), i = 1, 2, \dots, N $\\   ~ & $\left \langle w_i^2 \right \rangle =  \mu_i^2 + \sigma_i^2, i = 1, 2, \dots, N $\\   ~ & $\langle \alpha_i \rangle = \frac{2 c+1}{a +  \langle \beta \rangle \left \langle w_i^2 \right \rangle }, i = 1, 2, \dots, N$\\   
\hline 
\end{tabularx} 
\caption{The SVB-S scheme} 
\label{tab:VEM_student} 
\end{table*}  

\begin{table*}[t] 
\centering 
\newcolumntype{R}{>{\raggedleft\arraybackslash}X}%
\begin{tabularx}{1.2 \textwidth}{ |l|R| }  
\hline   
\multicolumn{2}{|l|}{\textbf{Sparse variational Bayes with a Laplace prior }} \\  
\hline  
Data likelihood & $p({\mathbf y}|{\mathbf w},\beta) = (2 \pi)^{-\frac{n}{2}} \beta^{\frac{n}{2}}  |\boldsymbol \Lambda|^{1/2} {\rm exp}\left[ - \frac{\beta}{2} \| \boldsymbol \Lambda ^{1/2}{\mathbf y} - \boldsymbol \Lambda ^{1/2} \mathbf{X} {\mathbf w}\|^2\right]$  \\ 
\hline  
Prior distributions  & $p(\beta;\rho,\delta) = \mathcal{G}(\beta;\rho,\delta) = \frac{\delta^\rho}{\Gamma(\rho)}\beta^{\rho-1} \mathrm{exp} \left[- \delta \beta \right]$ \\   ~ & $p({\mathbf w}|\boldsymbol \alpha,\beta) = \mathcal{N}(\mathbf{w}|0,\beta^{-1} \mathbf{A}^{-1}) = \prod_{i=1}^{N} \left(2\pi\right)^{-\frac{1}{2}} \beta ^{\frac{1}{2}} \alpha_i^\frac{1}{2} \mathrm{exp}\left[ - \frac{\beta}{2} w_i^2 \alpha_i \right]$  \\ 
~ & $p(\alpha_i|b) = \mathcal{IG}(\alpha_i|1,\frac{b}{2}) = \frac{b}{2}  \alpha_i^{-2}{\rm exp} \left[- \frac{b}{2} \frac{1}{\alpha_i} \right], i = 1, 2, \dots, N$ \\  
~ & $p(b;\kappa,\nu) = \mathcal{G}(b;\kappa,\nu) = \frac{\nu^\kappa}{\Gamma(\kappa)} b^{\kappa-1} \mathrm{exp} \left[- \nu b \right]$ \\  
\hline  
Variational distributions  & $q(\beta) =\mathcal{G}(\beta;\tilde{\rho},\tilde{\delta})$, with $\tilde{\rho} = \frac{n+N}{2} + \rho$ and $\tilde{\delta} = \delta +\frac{1}{2} \left \langle \| \boldsymbol \Lambda^{1/2} {\mathbf y} - \boldsymbol \Lambda^{1/2} \mathbf{X} {\mathbf w}\|^2\right \rangle + \frac{1}{2} \left  \langle  {\mathbf w} ^T {\mathbf A}{\mathbf w} \right \rangle$ \\   
~ & $q(w_i) = \mathcal{N}(w_i;\mu_i,\sigma_i^2)$, with $\sigma_i^2 = \langle \beta \rangle ^{-1} ( \mathbf{x}_i^T \boldsymbol \Lambda \mathbf{x}_i + \langle \alpha_i \rangle )^{-1}$ and $\mu_i = \langle  \beta \rangle \sigma_i^2 \mathbf{x}_i^T \boldsymbol \Lambda  ( {\mathbf y} - \mathbf{X}_{\neg i} \boldsymbol \mu_{\neg i }), i = 1, 2, \dots, N$\\   
~  & $q(\alpha_i) = \frac{\left ( \left \langle \beta \right \rangle \left \langle w_i^2 \right \rangle \slash \left \langle b \right \rangle \right)^{-1\slash 4}}{2 K_{1\slash 2} \left (\sqrt{\left \langle \beta \right \rangle \left \langle w_i^2 \right \rangle \left \langle b \right \rangle}\right)} \alpha_i^{-3\slash 2} \mathrm{exp}\left[ -  \frac{1}{2} \left \langle \beta \right \rangle \left \langle w_i^2 \right \rangle \alpha_i - \frac{\left \langle b \right \rangle}{2} \frac{1}{\alpha_i}\right], i = 1, 2, \dots, N$ \\   ~ & $q(b) = \mathcal{G}(b;\tilde \kappa,\tilde \nu)$, with $ \tilde \kappa = N+\kappa $, and $\tilde \nu = \nu + \frac{1}{2} \sum_{i=1}^N{ \left \langle \frac{1}{\alpha_i} \right \rangle } $ \\   
\hline   
Variational updates &  $\langle \beta \rangle = ( n+N + 2 \rho ) \slash (  2 \delta + \left \langle \|\boldsymbol \Lambda^{1/2}{\mathbf y} - \boldsymbol \Lambda^{1/2}\mathbf{X} \mathbf{w}\|^2 \right \rangle  + \sum_{i=1}^N \left \langle w_i^2 \right \rangle \langle \alpha_i \rangle 
 ) 
$ \\   ~ & $\sigma_i^2 = \langle \beta \rangle ^{-1} ( \mathbf{x}_i^T \boldsymbol \Lambda \mathbf{x}_i + \langle \alpha_i \rangle )^{-1}, i = 1, 2, \dots, N$ \\  
~ & $\langle w_i \rangle \equiv \mu_i = \langle  \beta \rangle \sigma_i^2 \mathbf{x}_i^T \boldsymbol \Lambda  ( {\mathbf y} - \mathbf{X}_{\neg i} \boldsymbol \mu_{\neg i }), i = 1, 2, \dots, N $\\   ~ & $\left \langle w_i^2 \right \rangle =  \mu_i^2 + \sigma_i^2, i = 1, 2, \dots, N $\\   ~ & $\langle \alpha_i \rangle = \sqrt{\frac{\left \langle b \right \rangle}{\left \langle \beta \right \rangle \left \langle w_i^2 \right \rangle }}, i = 1, 2, \dots, N$\\  
~ & $\left \langle \frac{1}{\alpha_i} \right \rangle = \gamma_i  = \frac{1}{\langle \alpha_i \rangle} + \frac{1}{\langle b \rangle}, i = 1, 2, \dots, N$ \\   ~ & $\langle b \rangle = \frac{N+\kappa}{\nu + \frac{1}{2} \sum_{i=1}^N{\left \langle \frac{1}{\alpha_i} \right \rangle} }$\\  
\hline
\end{tabularx} 
\caption{The SVB-L scheme} 
\label{tab:VEM_Laplace} 
\end{table*}

\begin{table*}[t] 
\centering
\newcolumntype{R}{>{\raggedleft\arraybackslash}X}%
\begin{tabularx}{1.2 \textwidth}{ |l|R| }  
\hline   
\multicolumn{2}{|l|}{\textbf{Sparse variational Bayes with a multiparameter Laplace prior }} \\ 
\hline   
Data likelihood & $p({\mathbf y}|{\mathbf w},\beta) = (2 \pi)^{-\frac{n}{2}} \beta^{\frac{n}{2}}  |\boldsymbol \Lambda|^{1/2} {\rm exp}\left[ - \frac{\beta}{2} \| \boldsymbol \Lambda ^{1/2}{\mathbf y} - \boldsymbol \Lambda ^{1/2} \mathbf{X} {\mathbf w}\|^2\right]$  \\  
\hline   Prior distributions  & $p(\beta;\rho,\delta) = \mathcal{G}(\beta;\rho,\delta) = \frac{\delta^\rho}{\Gamma(\rho)}\beta^{\rho-1} \mathrm{exp} \left[- \delta \beta \right]$ \\ 
~ & $p({\mathbf w}|\boldsymbol \alpha,\beta) = \mathcal{N}(\mathbf{w}|0,\beta^{-1} \mathbf{A}^{-1}) = \prod_{i=1}^{N} \left(2\pi\right)^{-\frac{1}{2}} \beta ^{\frac{1}{2}} \alpha_i^\frac{1}{2} \mathrm{exp}\left[ - \frac{\beta}{2} w_i^2 \alpha_i \right]$  \\ 
~ & $p(\alpha_i|b_i) = \mathcal{IG}(\alpha_i|1,\frac{b_i}{2}) = \frac{b_i}{2}  \alpha_i^{-2}{\rm exp} \left[- \frac{b_i}{2} \frac{1}{\alpha_i} \right], i = 1, 2, \dots, N$ \\ 
~ & $p(b_i;\kappa,\nu) = \mathcal{G}(b_i;\kappa,\nu) = \frac{\nu^\kappa}{\Gamma(\kappa)} b_i^{\kappa-1} \mathrm{exp} \left[- \nu b_i \right], i = 1, 2, \dots, N$ \\  
\hline 
Variational distributions  & $q(\beta) =\mathcal{G}(\beta;\tilde{\rho},\tilde{\delta})$, with $\tilde{\rho} = \frac{n+N}{2} + \rho$ and $\tilde{\delta} = \delta +\frac{1}{2} \left \langle \| \boldsymbol \Lambda^{1/2} {\mathbf y} - \boldsymbol \Lambda^{1/2} \mathbf{X} {\mathbf w}\|^2\right \rangle + \frac{1}{2} \left  \langle  {\mathbf w} ^T {\mathbf A}{\mathbf w} \right \rangle, i = 1, 2, \dots, N$ \\   ~ & $q(w_i) = \mathcal{N}(w_i;\mu_i,\sigma_i^2)$, with $\sigma_i^2 = \langle \beta \rangle ^{-1} ( \mathbf{x}_i^T \boldsymbol \Lambda \mathbf{x}_i + \langle \alpha_i \rangle )^{-1}$ and $\mu_i = \langle  \beta \rangle \sigma_i^2 \mathbf{x}_i^T \boldsymbol \Lambda  ( {\mathbf y} - \mathbf{X}_{\neg i} \boldsymbol \mu_{\neg i }), i = 1, 2, \dots, N$\\   
~  & $q(\alpha_i) = \frac{\left ( \left \langle \beta \right \rangle \left \langle w_i^2 \right \rangle \slash \left \langle b_i \right \rangle \right)^{-1\slash 4}}{2 K_{1\slash 2} \left (\sqrt{\left \langle \beta \right \rangle \left \langle w_i^2 \right \rangle \left \langle b_i \right \rangle}\right)} \alpha_i^{-3\slash 2} \mathrm{exp}\left[ -  \frac{1}{2} \left \langle \beta \right \rangle \left \langle w_i^2 \right \rangle \alpha_i - \frac{\left \langle b_i \right \rangle}{2} \frac{1}{\alpha_i}\right], i = 1, 2, \dots, N$ \\  
~ & $q(b_i) = \mathcal{G}(b_i;\tilde \kappa,\tilde \nu )$, with $\tilde \kappa = 1 + \kappa$ and $\tilde \nu = \nu + \frac{1}{2} \left \langle \frac{1}{\alpha_i} \right \rangle, i = 1, 2, \dots, N $ \\   
\hline   
Variational updates &  $\langle \beta \rangle = ( n+N  + 2 \rho ) \slash ( 2 \delta + \left \langle \|\boldsymbol \Lambda^{1/2}{\mathbf y} - \boldsymbol \Lambda^{1/2}\mathbf{X} \mathbf{w}\|^2 \right \rangle  +  \sum_{i=1}^N \left \langle w_i^2 \right \rangle  \langle \alpha_i \rangle  )$ \\  
~ & $\sigma_i^2 = \langle \beta \rangle ^{-1} ( \mathbf{x}_i^T \boldsymbol \Lambda \mathbf{x}_i + \langle \alpha_i \rangle )^{-1}, i = 1, 2, \dots, N$ \\ 
~ & $\langle w_i \rangle \equiv \mu_i = \langle  \beta \rangle \sigma_i^2 \mathbf{x}_i^T \boldsymbol \Lambda  ( {\mathbf y} - \mathbf{X}_{\neg i} \boldsymbol \mu_{\neg i }), i = 1, 2, \dots, N $\\   ~ & $\left \langle w_i^2 \right \rangle =  \mu_i^2 + \sigma_i^2, i = 1, 2, \dots, N $\\ 
~ & $\langle \alpha_i \rangle = \sqrt{\frac{\left \langle b_i \right \rangle}{\left \langle \beta \right \rangle \left \langle w_i^2 \right \rangle }}, i = 1, 2, \dots, N$\\ 
~ & $\left \langle \frac{1}{\alpha_i} \right \rangle = \gamma_i  = \frac{1}{\langle \alpha_i \rangle} + \frac{1}{\langle b_i \rangle}, i = 1, 2, \dots, N$ \\   ~ & $\langle b_i \rangle = \frac{1+\kappa}{\nu + \frac{1}{2 } \left \langle \frac{1}{\alpha_i} \right \rangle }, i = 1, 2, \dots, N$\\  
\hline 
\end{tabularx} 
\caption{The SVB-mpL scheme} 
\label{tab:VEM_multiLaplace} 
\end{table*}

\section{Sparse variational Bayes Adaptive Estimation} 
\label{sec:sbl_ad_filtering}

The variational schemes presented in Tables \ref{tab:VEM_student}, \ref{tab:VEM_Laplace} and \ref{tab:VEM_multiLaplace} deal with the batch estimation problem associated with (\ref{eq:rls_cost_function}), that is, given the $n\times N$ data matrix ${\mathbf X}$ and the $n\times 1$ vector of observations ${\mathbf y}$, they provide a sparse estimate ($\hat{\mathbf w} \equiv \boldsymbol \mu$) of ${\mathbf w}$ after a few iterations. However, in an adaptive estimation setting, solving the size-increasing (by $n$) batch problem in each time iteration is computationally prohibitive. Therefore, SVB-S, SVB-L and SVB-mpL should be properly modified and adjusted in order to perform adaptive processing in a computationally efficient manner, giving rise to ASVB-S, ASVB-L and ASVB-mpL respectively. In this regard, the time index $n$ is reestablished here and the expectation operator $\langle \cdot \rangle$ is removed from the respective parameters, keeping in mind that henceforth these will refer to {\it posterior} 
parameters. 
By carefully inspecting (\ref{eq:sigmai}), (\ref{eq:mui}), (\ref{eq:beta_exp}), and (\ref{eq:beta_exp_den}) (which are common for all three schemes) we reveal the following time-dependent quantities that are commonly met in LS estimation tasks,  
\begin{align}  
& \mathbf{R}(n) = \mathbf{X}^T(n) \boldsymbol \Lambda(n) \mathbf{X}(n) + \mathbf{A} (n-1), \label{eq:autcor}\\  
& \mathbf{z}(n) = \mathbf{X}^T(n) \boldsymbol \Lambda(n) \mathbf{y}(n), \label{eq:crosscor} \\  
& d(n) = \mathbf{y}^T(n) \boldsymbol \Lambda(n) \mathbf{y}(n). \label{eq: energ} 
\end{align} 
Note that in (\ref{eq:autcor}) a time-delayed regularization term $\mathbf{A}(n-1)$ is considered. This is related to the update ordering of the various algorithmic quantities and does affect the derivation and performance of the new algorithms. From the definitions of $\mathbf{y}(n)$ and $\mathbf{X}(n)$ in (\ref{eq:yn}) and (\ref{eq:Xn}) and that of $\boldsymbol \Lambda(n)$, it is easily shown that $\mathbf{R}(n), \mathbf{z}(n)$ and $d(n)$ can be efficiently time-updated as follows: 
\begin{alignat}{3} 
&{\mathbf R} (n) &&= \lambda {\mathbf R} (n-1) +  {\mathbf x}(n) {\mathbf x}^T(n) - \lambda \mathbf{A} (n-2) +  \mathbf{A}(n-1) \label{eq:Sigma_inv_time_upd} \\ 
&{\mathbf z} (n) &&= \lambda  {\mathbf z} (n-1) + {\mathbf x}(n)  y(n) \label{eq:z_time_upd}, \\ 
&d(n) &&= \lambda d(n-1) + y^2(n). \label{eq:d_time_upd} 
\end{alignat} 
It is readily recognized that $\mathbf{R}(n)$ is the exponentially weighted sample autocorrelation matrix of ${\mathbf x}(n)$ regularized by the diagonal matrix $\mathbf{A} (n-1)$, $\mathbf{z}(n)$ is the exponentially weighted crosscorrelation vector between ${\mathbf x}(n)$ and $y(n)$, and $d(n)$ is the exponentially weighted energy of the observation vector ${\mathbf y}(n)$. By substituting (\ref{eq:sigmai}) in  (\ref{eq:mui}) (with the time index $n$ now included) and using (\ref{eq:autcor}) and (\ref{eq:crosscor}), it is straightforward to show that the adaptive weights $\hat{w}_i(n) (\equiv \mu_i(n))$ can be efficiently computed in time for $i=1,2,\ldots,N$, as follows 
\begin{align} 
\hat{w}_i(n) = \frac{1}{r_{ii}(n)} \left( z_i(n) - \mathbf{r}_{\neg i}^T(n) \hat{\mathbf{w}}_{\neg i}(n) \right). 
\label{eq:w_i_update} 
\end{align} 
In the last equation, $z_i(n) = \mathbf{x}_i^T(n)\boldsymbol \Lambda (n) \mathbf{y}(n)$ is the $i$-th element of $\mathbf{z}(n)$, $r_{ii}(n) = \mathbf{x}_i^T(n)\boldsymbol \Lambda (n) \mathbf{x}_i(n) + \alpha_i(n-1)$ is the $i$-th diagonal element of $\mathbf{R}(n)$, $\mathbf{r}^T_{\neg i}(n) = \mathbf{x}_i^T(n)\boldsymbol \Lambda (n) \mathbf{X}_{\neg i}(n)$ is the $i$-th row of $\mathbf{R}(n)$ after removing its $i$-th element $r_{ii}(n)$, and  
\begin{align} 
\hat{\mathbf{w}}_{\neg i}(n) = [\hat{w}_1(n), \dots, \hat{w}_{i-1}(n), \hat{w}_{i+1}(n-1), \dots, \hat{w}_N(n-1)]^T.  
\label{eq:wni} 
\end{align} 
From (\ref{eq:w_i_update}) and (\ref{eq:wni}) it is easily noticed that each weight estimate $\hat{w}_i(n)$ depends on the most recent estimates in time of the other $N-1$ weights. This is in full agreement with the spirit of the variational Bayes approach and the batch SVB schemes presented in the previous Section, where each model parameter is computed based on the most recent values of the remaining parameters. As far as the noise precision parameter $\beta (n)$ is concerned, despite its relatively complex expression given in (\ref{eq:beta_exp}), it is shown in Appendix \ref{app:betan} that it can be estimated with $\mathcal{O}(N)$ operations per time iteration as follows  
\begin{align}  
\beta(n) = \frac{(1-\lambda)^{-1} + N +2\rho}{2\delta + d(n) - \mathbf{z}^T(n) \hat{\mathbf{w}}(n-1) +\mathbf{r}^{T}(n)\boldsymbol \sigma(n-1)  }. 
\label{eq:beta_update}  
\end{align} 
In (\ref{eq:beta_update}), the term $(1-\lambda)^{-1}$ represents the active time window in an exponentially weighted LS setting, $\mathbf{r}(n) = \mathrm{diag}(\mathbf{R}(n))$ and $\boldsymbol \sigma(n-1) = [\sigma_1^2(n-1),\sigma_2^2(n-1),\ldots,\sigma_N^2(n-1) ]^T$ is the vector of posterior weight variances at time $n-1$ with 
\begin{align} 
\sigma_i^2(n-1) = \frac{1}{\beta (n-1) r_{ii}(n-1)}, \label{eq:wivar} 
\end{align} 
according  to Eq. (\ref{eq:sigmai}). Note that Eqs. (\ref{eq:w_i_update}) and (\ref{eq:beta_update}) are common in all adaptive schemes described in this paper. What differentiates the algorithms is the way their sparsity enforcing precision parameters $\alpha_i(n)$ are computed in time. More specifically, from (\ref{eq:alpha_i_exp}), (\ref{eq:sigmai}) and the fact that $\langle w_i^2 \rangle = \hat{w}_i^2 + \sigma_i^2$, we get for ASVB-S,  
\begin{align} 
\alpha_i(n) = \frac{2c+1}{a +\beta(n)\hat{w}_i^2(n) + r^{-1}_{ii}(n)}. 
\label{eq:aiup} 
\end{align} 
Concerning ASVB-L, from Table \ref{tab:VEM_Laplace} we obtain the following time update recursions, 
\begin{align} 
\alpha_i(n) = \sqrt{\frac{b(n-1)}{\beta(n)\hat{w}_i^2(n) + r^{-1}_{ii}(n)}} \label{eq:aiupL} \end{align}  \begin{align} \gamma_i(n) = \frac{1}{\alpha_i(n)} + \frac{1}{b(n-1)} 
\label{eq:gammaiL} 
\end{align}
\begin{align}
b(n) = \frac{N+\kappa}{\nu + \frac{1}{2}\sum_{i=1}^{N}\gamma_i(n)}.
\label{eq:bL} 
\end{align} 
Finally, for ASVB-mpL we get expressions similar to (\ref{eq:aiupL}) and (\ref{eq:gammaiL}) with $b(n-1)$ being replaced by $b_i(n-1)$, while $b_i(n)$ is expressed as  
\begin{align}
b_i(n) = \frac{1+\kappa}{\nu + \frac{1}{2}\gamma_i(n)} 
\label{eq:biL}
\end{align}
The main steps of the proposed adaptive sparse variational Bayes algorithms are given in Table \ref{tab:prop_alg}. In the Table, the hyperparameters $\rho,\delta,\kappa$ and $\nu$ take very small values (of the order of $10^{-6}$). All three algorithms have robust performance, which could be attributed to the absence of matrix inversions or other numerically sensitive computation steps. The algorithms are based on second-order statistics and have an $\mathcal{O}(N^2)$ complexity, similar to that of the classical RLS. Their most computationally costly steps, which require $\mathcal{O}(N^2)$ operations, are those related to the updates of $\mathbf R (n)$ and $\hat{\mathbf w}(n)$. Note, though, that in an adaptive {\it filtering} setting, this complexity can be dramatically reduced (and become practically $\mathcal{O}(N)$) by taking advantage of the underlying shift invariance 
property of the data vector $\mathbf{x}(n)$ \cite{2010_angelosante}. As shown in the simulations of Section \ref{sec:experiments}, the algorithms converge very fast to sparse estimates for $\mathbf w(n)$ and in the case of ASVB-S and ASVB-mpL, offer lower steady-state estimation error compared to other competing deterministic sparse adaptive schemes. Additionally, while the latter require knowledge of the noise variance beforehand, this variance is naturally estimated in time as $1/\beta(n)$ during the execution of the new algorithms. 

Most recently reported deterministic sparse adaptive estimation algorithms are sequential variants of the lasso estimator, performing variable selection via soft-thresholding, e.g. the algorithms developed in \cite{2010_angelosante}. To achieve their best possible performances though, such approaches necessitate the use of suitably selected regularization parameters, whose values, in most cases, are determined via time-demanding cross-validation and fine-tuning. Moreover, this procedure should be repeated depending on the application and the application conditions. Unlike the approach followed in deterministic schemes, a completely different sparsity inducing mechanism is used in the proposed algorithms. More specifically, as the algorithms progress in time, many of the exponentially distributed precision parameters $\alpha_i(n-1)$'s are automatically driven to very large values, forcing also the corresponding diagonal elements $r_{ii}(n)$ of $\mathbf R (n)$ to become excessively large (Eq. (\ref{eq:autcor}))
. 
As a result, according to (\ref{eq:w_i_update}), many weight parameters are forced to become almost zero, thus imposing sparsity. Notably, this sparsity inducing mechanism alleviates the need for fine-tuning or cross-validating of any parameters, which makes the proposed schemes fully automated, and thus, particularly attractive from a practical point of view.     %

\begin{table}[t] 
\centering
\begin{tabular}{ |p{90mm}|} 
\hline \\ [-1.5ex]
Initialize $\lambda, \hat{ \mathbf w}(0), \mathbf A (-1), \mathbf A (0), \mathbf{R}(0), \mathbf z (0), d (0), \boldsymbol \sigma (0)$ \\ 
Set $c, a, \rho, \delta, \kappa, \nu $ to very small values \\ 
\textbf{for} $n = 1, 2, \dots$ \\
\quad ${\mathbf R} (n) = \lambda {\mathbf R} (n-1) +  {\mathbf x}(n) {\mathbf x}^T(n) - \lambda \mathbf{A} (n-2) +  \mathbf{A}(n-1)$ \\ 
\quad ${\mathbf z} (n) = \lambda  {\mathbf z} (n-1) + {\mathbf x}(n)  y(n)$ \\ \quad $d(n) = \lambda d(n-1) + y^2(n)$ \\ 
\quad $\beta(n) = \frac{N +(1-\lambda)^{-1}+2\rho }{ 2\delta + d(n) - \mathbf{z}^T(n) \hat{\mathbf{w}}(n-1) +\mathbf{r}^{T}(n)\boldsymbol \sigma (n-1)}$ \\ 
\quad \textbf{for} $i = 1, 2, \dots, N$ \\
\quad \quad $\sigma_i^2(n) = 1 \slash (\beta(n)r_{ii}(n))$\\ 
\quad \quad $\hat{w}_i(n) = r_{ii}^{-1}(n) \left( z_i(n) - \mathbf{r}_{\neg i}^T(n) \hat{\mathbf{w}}_{\neg i}(n) \right)$\\
\quad \quad \fbox{\parbox{0.8\linewidth}{ \textbf{ASVB-S}\\ $ \alpha_i(n) = \left (  2 c + 1 \right) \slash \left ( a + \beta(n)\hat{w}_i^2(n)  + r_{ii}^{-1}(n) \right)$ }}\\ [2.7ex]
\quad \quad \fbox{\parbox{0.8\linewidth}{ \textbf{ASVB-L}\\ $\alpha_i(n) =  \sqrt { b(n-1) \slash  (  \beta(n)\hat{w}_i^2(n)  + r_{ii}^{-1}(n)  ) }$ \\ $\gamma_i(n) =  1 \slash \alpha_i(n) + 1 \slash b(n-1)$ }}\\[4ex]
\quad \quad \fbox{\parbox{0.8\linewidth}{ \textbf{ASVB-mpL}\\ $\alpha_i(n) =  \sqrt { b_i(n-1) \slash  (  \beta(n)\hat{w}_i^2(n)  + r_{ii}^{-1}(n)  ) }$ \\ $\gamma_i(n) =  1 \slash \alpha_i(n) + 1 \slash b_i(n-1)$ \\ $b_i(n) = (1 + \kappa) \slash \left( \nu +  \gamma_i(n) \slash 2 \right)$}}\\ [5ex]
%
\quad \textbf{end for} \\ 
\quad \fbox{\parbox{0.8\linewidth}{ \textbf{ASVB-L}\\ $b(n) = (N + \kappa) \slash \left( \nu + \frac{1}{2} \sum_{i=1}^N  \gamma_i(n)\right)$}}\\ [3ex]
\textbf{end for} \\
\hline 
\end{tabular} 
\caption{The proposed ASVB-S, ASVB-L, and ASVB-mpL algorithms} 
\label{tab:prop_alg} 
\end{table}

\section{Discussion on the proposed algorithms} 
\label{sec:analysis} 

Let us now concentrate on the weight updating mechanism given in (\ref{eq:w_i_update}), which is common in all proposed schemes, and attempt to interpret it. To this end, we define the following regularized LS cost function, 
\begin{align} 
{\cal J}_\mathrm{LS-R}(n) & = \| \boldsymbol \Lambda^{1/2}(n) {\mathbf y}(n) -\boldsymbol \Lambda^{1/2}(n) {\mathbf X}(n) \hat {\mathbf w}(n)\|^2 \nonumber \\ 
& + \hat {\mathbf w}^T(n){\mathbf A}(n-1) \hat {\mathbf w}(n), 
\label{eq:rlscf}
\end{align}
where the diagonal matrix ${\mathbf A}(n-1)$ has positive diagonal entries and is assumed known, (i.e. for the moment we ignore the procedure that produces ${\mathbf A}(n-1)$). As it is well-known the vector $\hat {\mathbf w}(n)$ that minimizes ${\cal J}_\mathrm{LS-R}(n)$ is the solution of the following system of equations,
\begin{align}
{\mathbf R}(n)\hat {\mathbf w}(n) = {\mathbf z}(n) 
\label{eq:regls} 
\end{align} 
where ${\mathbf R}(n)$ and ${\mathbf z}(n)$ are given in (\ref{eq:autcor}) and (\ref{eq:crosscor}), respectively. Let us now write, 
\begin{align} 
{\mathbf R}(n) = {\mathbf L}(n)+{\mathbf D}(n)+{\mathbf U}(n),  
\label{eq:gsfact} 
\end{align} 
where ${\mathbf L}(n)$ is the strictly lower triangular component of ${\mathbf R}(n)$, ${\mathbf D}(n)$ its diagonal component and ${\mathbf U}(n)$ its strictly upper triangular component. This matrix decomposition is the basis of the Gauss-Seidel method \cite{1996_golub},  and, if substituted in (\ref{eq:regls}) leads to the following iterative scheme for obtaining the optimum $\hat {\mathbf w}(n)$, 
\begin{align} 
({\mathbf D}(n)+ {\mathbf L}(n))\hat {\mathbf w}^{(k)}(n) = {\mathbf z}(n) - {\mathbf U}(n)\hat {\mathbf w}^{(k-1)}(n) ,
\label{eq:gausseidel} 
\end{align} 
where $k$ is the iterations index and the time index $n$ is considered fixed (batch mode). 
From the last equation, it is easily verified that by using forward substitution, the elements of ${\mathbf w}^{(k)}(n)$ can be computed sequentially as follows for $i = 1, 2, \dots, N$, 
\begin{align} 
\hat{w}_i^{(k)}(n)=\frac{1}{r_{ii}(n)} \left( z_i(n) - \sum_{j<i}r_{ij}(n)\hat{w}_j^{(k)}(n) - \sum_{j>i}r_{ij}(n)\hat{w}_j^{(k-1)}(n) \right ).
\label{eq:gsel} 
\end{align} 
Since the regularized autocorrelation matrix ${\mathbf R}(n)$ is symmetric and positive definite, the Gauss-Seidel scheme in (\ref{eq:gausseidel}) converges (for $n$ fixed) after a few iterations to the solution of (\ref{eq:regls}), irrespective of the initial choice for $\hat {\mathbf w}^{(0)}(n)$ \cite{1996_golub}. Therefore, in an adaptive estimation setting, optimization is achieved by executing a sufficiently high number of Gauss-Seidel iterations in each time step $n$. An alternative, more computationally efficient approach though, is to match the iteration and time indices, $k$ and $n$ in (\ref{eq:gsel}); that is to consider that a single time iteration $n$ of the adaptive algorithm entails just a {\it single} iteration of the Gauss-Seidel procedure over each coordinate of $\hat {\mathbf w}(n)$. By doing so, we end up with the weight updating formula given previously in (\ref{eq:w_i_update}). Such a Gauss-Seidel adaptive algorithm has been previously reported in \cite{1999_xu,2004_bose} for the 
conventional LS cost function ${\cal J}_\mathrm{LS}(n)$ given in (\ref{eq:rls_cost_function}), without considering any regularization and/or sparsity issues. It has been termed as the \emph{Euclidean direction set} (EDS) algorithm. Relevant convergence results have been also presented in \cite{1998_xu}. However, in that analysis the {\it time-invariant} limiting values of the autocorrelation and crosscorelation quantities have been employed and thus, the obtained convergence results are not valid for the adaptive Gauss-Seidel algorithm described in \cite{1999_xu,2004_bose}.

Apart from the Gauss-Seidel viewpoint presented above, a different equivalent approach to arrive at the same weight updating formula as in (\ref{eq:w_i_update}) is the following. We start with the cost function in (\ref{eq:rlscf}) and minimize it w.r.t. a single weight component in a cyclic fashion. This leads to a cyclic coordinate descent (CCD) algorithm \cite{1999_bertsekas} for minimizing ${\cal J}_\mathrm{LS-R}(n)$ for $n$ fixed. If we now execute only one cycle of the CCD algorithm per time iteration $n$, we obtain an adaptive algorithm whose weight updating formula is expressed as in (\ref{eq:w_i_update}).  CCD algorithms for {\it sparse adaptive} estimation have been recently proposed in \cite{2010_angelosante}. These algorithms, however, are based on the minimization of ${\cal J}_{\mathrm{LS-}\ell_1}(n)$ given in (\ref{eq:rls_l1__cost_function}), which explicitly incorporates an $\ell_1$ penalizing term. In \cite{2010_angelosante} the proposed algorithms have been supported theoretically by 
relevant convergence results. To the best of our knowledge, \cite{2010_angelosante} is the only contribution where a proof of convergence of CCD adaptive algorithms has been presented and documented. 

From the previous analysis, we conclude that the proposed fully factorized variational methodology described in this paper leads to adaptive estimation schemes where, a) the model weights are adapted in time by using a Gauss-Seidel or CCD type updating rule and b)  explicit mechanisms (different for each algorithm) are embedded for computing in time the regularization matrix ${\mathbf A}(n)$ that imposes sparsity to the adaptive weights. The algorithms are fully automated, alleviating the need for predetermining and/or fine-tuning of any penalizing or other regularization parameters.

The convergence properties of the proposed algorithmic family is undoubtedly of major importance. Notice that in \cite{2010_angelosante} the following ergodicity assumptions are made as a prerequesite for proving convergence,  
\begin{align} 
\lim_{n\rightarrow\infty} \mathrm{ Prob} \left[  \frac{1}{n}{\mathbf R}(n)={\mathbf R}_{\infty} \right ] =1 \mathrm{\ and\ } {\mathbf R}_{\infty} \mathrm{ positive \ definite}  
\label{eq:ergR}
\end{align} 
\begin{align} 
\lim_{n\rightarrow\infty} \mathrm{ Prob} \left[   \frac{1}{n}{\mathbf z}(n)={\mathbf z}_{\infty} \right ] =1 ,
\label{eq:ergz}
\end{align}
where  ${\mathbf R}(n) = {\mathbf X}^T(n){\mathbf X}(n)$ and ${\mathbf z}(n) = {\mathbf X}^T(n){\mathbf y}(n)$ in \cite{2010_angelosante}. If these assumptions hold in our case (with ${\mathbf R}(n)$ defined as in (\ref{eq:autcor})) then the convergence analysis presented in \cite{2010_angelosante} would be also valid for the adaptive algorithms described in this paper, with only slight modifications. For this to happen, matrix ${\mathbf A}(n)$ should be either constant, or dependent solely on the data. This is, however, not true owing to the nonlinear dependence of $a_i(n)$'s on the  corresponding weight components as shown in (\ref{eq:aiup}) and (\ref{eq:aiupL}). Such  a nonlinear interrelation among the parameters of the adaptive algorithms renders the analysis of their convergence an extremely difficult task. In any case, relevant efforts have been undertaken and the problem is under current investigation.

\section{Experimental results}
\label{sec:experiments}

In this section we present experimental results obtained from applying the proposed variational algorithmic framework to the estimation of a time-varying sparse wireless channel. To assess the estimation performance of the proposed adaptive sparse variational Bayesian algorithms\footnote{A Matlab implementation of the variational framework presented in this paper is publicly available at http://members.noa.gr/themelis.}, a comparison against a number of state-of-the-art deterministic adaptive algorithms is made, such as the sparsity agnostic RLS, \cite{2001_haykin}, the sparse RLS (SPARLS), \cite{2010_babadi}, the time weighted lasso (TWL), \cite{2010_angelosante}, and the time and norm weighted lasso (TNWL), \cite{2010_angelosante}. Moreover, an RLS that operates only on the {\it a priori} known support set of the channel coefficients, termed as the genie aided RLS (GARLS), is also included in the experiments, in order to serve as a benchmark. 
To set a fair comparison from a performance point of view, the optimal parameters of the deterministic algorithms are obtained via exhaustive cross-validation in order to acquire the best of their performances.


We consider a wireless channel with $64$ coefficients, which are generated according to Jake's model, \cite{1994_jakes}. Unless otherwise stated, only $8$ of these coefficients are nonzero, having arbitrary positions (support set), and following a Rayleigh distribution with normalized Doppler frequency $f_d\ T_s = 5\times10^{-5}$. The forgetting factor is set to $\lambda = 0.99$. The channel's input is a random sequence of binary phase-shift keying (BPSK) $\pm 1$ symbols. The symbols are organized in packets of length $1000$ per transmission. Gaussian noise is added to the channel, whose variance is adjusted according to the SNR level of each experiment. The estimation performance of the algorithms is measured in terms of the normalized mean square error (NMSE), which is defined as 
\begin{align}
 \mathrm{NMSE} =  \frac{ \left \langle   \|\mathbf{w} - \hat{\mathbf{w}}\|^2 \right \rangle  }{ \left \langle  \|\mathbf{w}\|^2 \right \rangle  }, 
 \label{eq:mse_definition}
\end{align}
where $\hat{ \mathbf w }$ is the estimate of the actual channel vector $\mathbf w$. All performance curves are ensemble average of $200$ transmission packets, channels, and noise realizations.

\begin{figure}[!t]
\centering
\includegraphics[width=0.9 \linewidth]{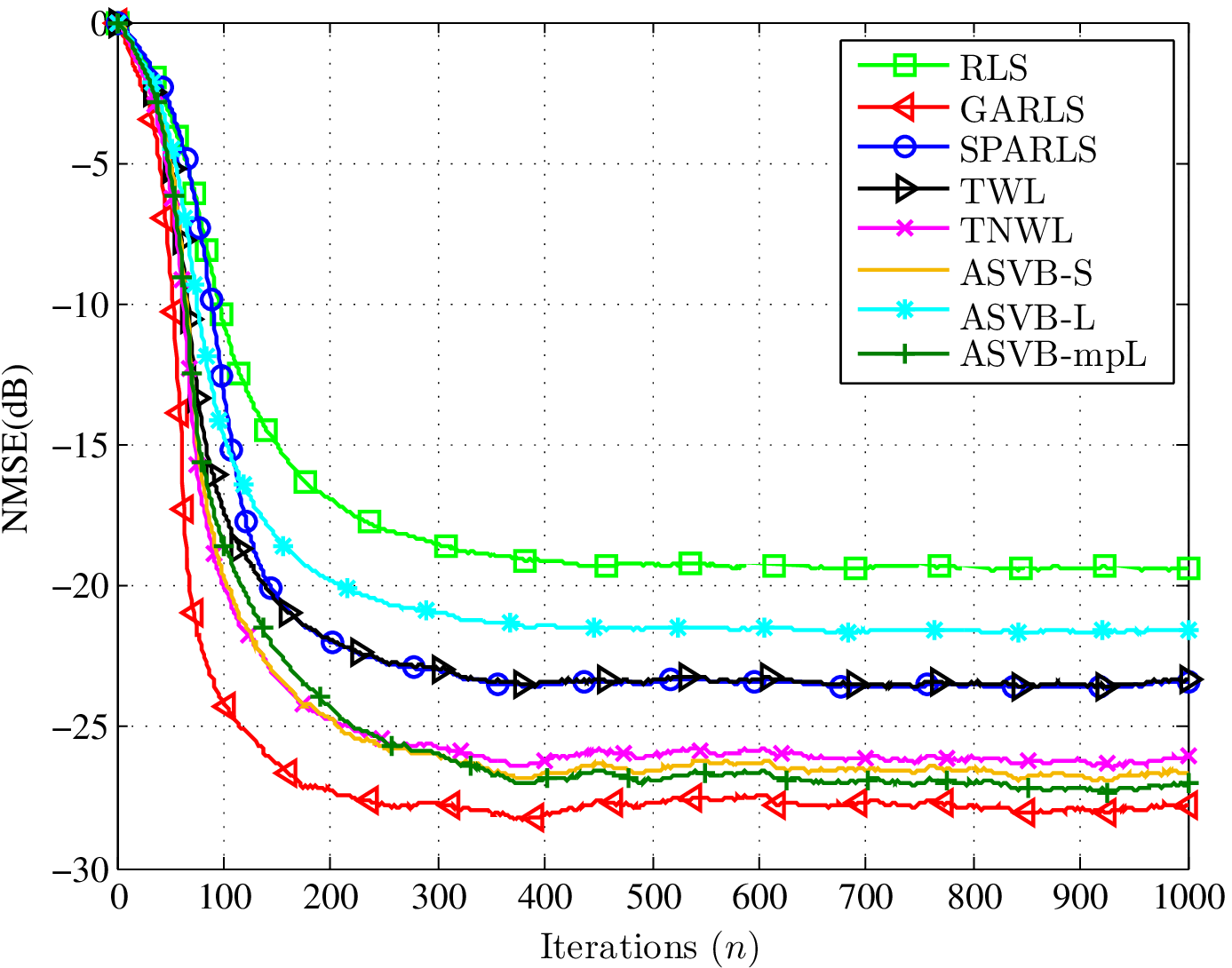}
\caption{NMSE curves of adaptive algorithms applied to the estimation of a sparse $64$-length time-varying channel with $8$ nonzero coefficients. The SNR is set to $15$dB. }
\label{fig:asbl_mse_vs_iteration}
\end{figure}

The first experiment demonstrates the estimation performance of the sparse adaptive estimation algorithms. Fig. \ref{fig:asbl_mse_vs_iteration} shows the NMSE curves of the RLS, GARLS, SPARLS, TWL, TNWL, ASVB-S, ASVB-L, and ASVB-mpL versus time. The SNR is set to $15$dB. Observe that all sparsity aware algorithms perform better than the RLS algorithm, whose channel tap estimates always take non-zero values, even if the actual channel coefficients are zero. Interestingly, there is an improvement margin of about $8$dB in the steady-state NMSE between the RLS and the GARLS, which, as expected, achieves the overall best performance. Moreover, the proposed ASVB-L algorithm has better performance than RLS, but although it promotes sparse estimates, it does not reach the performance level of ASVB-S and ASVB-mpL. From Fig. \ref{fig:asbl_mse_vs_iteration} it is clear that both ASVB-S and ASVB-mpL outperform TNWL, which, in turn, has the best performance among the deterministic algorithms. The ASVB-mpL algorithm 
reaches 
an error floor that is closer to the one of GARLS, and it provides an NMSE improvement of $1$dB over TNWL and $3$dB over SPARLS and TWL.

At this point we should shed some light on the relationship between the estimation performance and the complexity of the deterministic algorithms. In a nutshell, the key objective of SPARLS, TWL and TNWL is to optimize the $\ell_1$ regularized LS cost function given in (\ref{eq:rls_l1__cost_function}) w.r.t. $\hat{\mathbf w}(n)$ and in a sequential manner. Their estimates, however, are inherently sensitive to the selection of the sparsity imposing parameter $\tau$. The NMSE curves shown in Fig. \ref{fig:asbl_mse_vs_iteration} are obtained after fine-tuning the values of the respective parameters of SPARLS, TWL and TNWL through extensive experimentation. Nonetheless, the thus obtained gain in estimation accuracy adds to the computational complexity of the optimization task. On the other hand, the proposed adaptive variational methods are fully automatic, all parameters are directly inferred from the data, and a single execution suffices to provide the depicted experimental results.

Observe also in Fig \ref{fig:asbl_mse_vs_iteration} that, as expected, all sparsity aware algorithms converge faster than RLS, requiring an average of approximately $100$ fewer iterations in order to reach the NMSE level of $-17$dB compared to RLS. Among the deterministic algorithms, TNWL is the one with the fastest convergence rate. In comparison, ASVB-mpL needs almost $10$ iterations more than TNWL to converge, but it converges to a lower error floor. Again, the convergence speed of the GARLS is unrivaled.

\begin{figure}[!t]
\centering
\includegraphics[width=0.9 \linewidth]{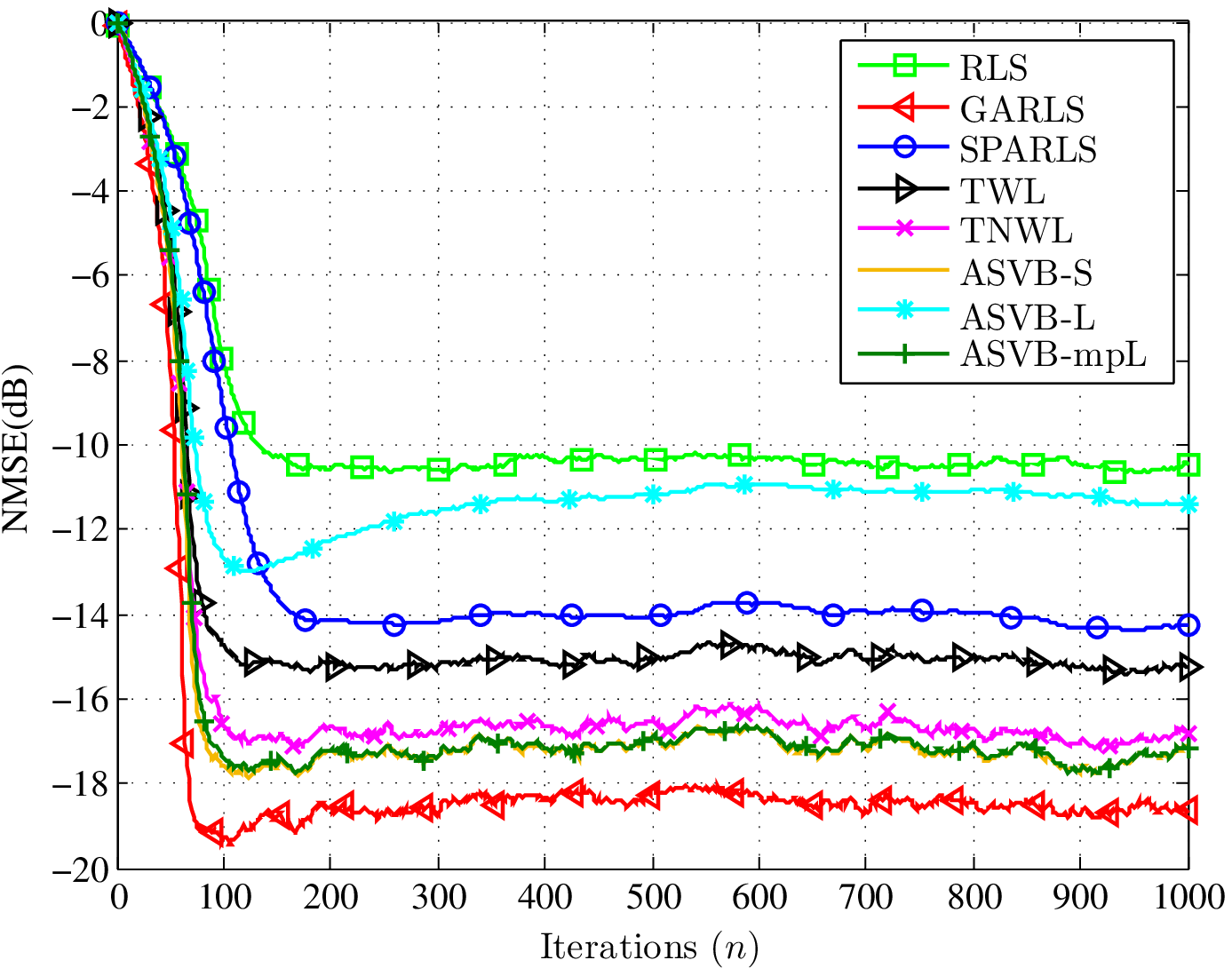}
\caption{NMSE curves of adaptive algorithms applied to the estimation of a fast-fading sparse $64$-length time-varying channel with $8$ nonzero coefficients. The SNR is set to $15$dB.}
\label{fig:asbl_mse_vs_iteration_fast}
\end{figure}

The next experiment explores the performance of the proposed algorithms for a fast fading channel. The settings of the first experiment are kept the same, with the difference that the normalized Doppler frequency is now increased to $f_d\ T_s = 8.35\times 10^{-4}$, that suits better to a high mobility application. Specifically, this Doppler results for a system operating at a carrier frequency equal to $1.8$GHz, with a sampling period $T_{s}=5\times10^{-6}$ and a mobile user velocity $100$Km/h. To account for fast channel variations, the forgetting factor is reduced to $\lambda = 0.96$ (except for ASVB-L, where $\lambda = 0.98$ is used). 
Fig. \ref{fig:asbl_mse_vs_iteration_fast} shows the resulting NMSE curves for all algorithms versus the number of iterations. In comparison to Fig. \ref{fig:asbl_mse_vs_iteration}, we observe that the steady-state NMSE of all algorithms has an expected increase. The algorithms' relative performance is the same, with the exception of ASVB-L, which has higher relative steady-state NMSE and is sensitive to $\lambda$. 
Nevertheless, the proposed ASVB-S and ASVB-mpL converge to a better error floor compared to all deterministic algorithms and their NMSE margin to TNWL is more perceptible now. 

\begin{figure}[!t]
\centering
\includegraphics[width=0.9 \linewidth]{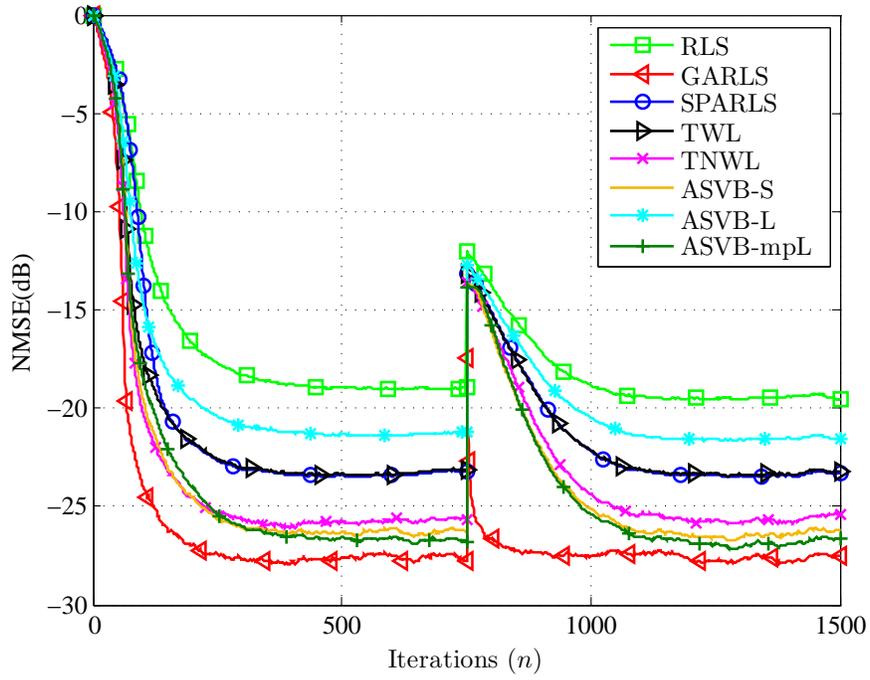}
\caption{NMSE curves of adaptive algorithms applied to the estimation of a sparse $64$-length time-varying channel with $8$ nonzero coefficients, with a non-zero coefficient added at the $750$th time mark. The SNR is set to $15$dB.}
\label{fig:asbl_mse_vs_iteration_spike}
\end{figure}

In the next simulation example, we investigate the tracking performance of the proposed sparse variational algorithms. The experimental settings are identical to those of Fig. \ref{fig:asbl_mse_vs_iteration}, with the exception that the packet length is now increased to $1500$ symbols, and an extra non-zero Rayleigh fading coefficient is added to the channel at the $750$th time instant. Note that until the $750$th time mark all algorithms have converged to their steady state. The resulting NMSE curves versus time are depicted in Fig. \ref{fig:asbl_mse_vs_iteration_spike}. The abrupt change of the channel causes all algorithms to record a sudden fluctuation in their NMSE curves. Nonetheless, the proposed ASVB-S and ASVB-mpL respond faster than the other algorithms to the sudden change and they successfully track the channel coefficients until they converge to error floors that are again closer to the benchmark GARLS.

\begin{figure}[!t]
\centering
\includegraphics[width=0.9 \linewidth]{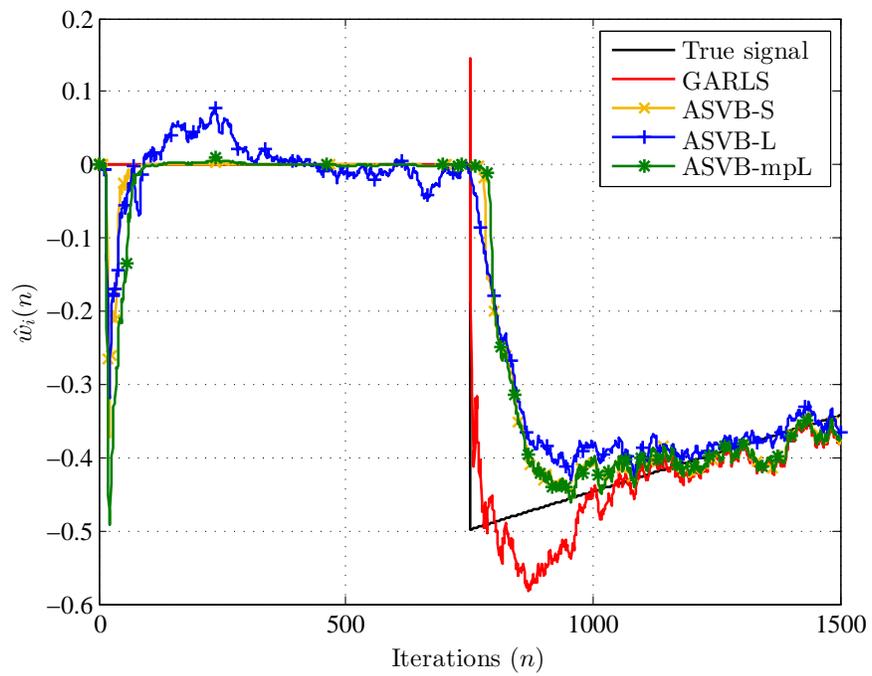}
\caption{Tracking of a time-varying channel coefficient.}
\label{fig:tap_vs_iteration}
\end{figure}

To get a closer look, Fig. \ref{fig:tap_vs_iteration} depicts the variations in time of the added channel coefficient and the respective estimates of the proposed algorithms. Notice by Fig. \ref{fig:tap_vs_iteration} that after the first $100$ iterations the ASVB-S and ASVB-mpL have converged to a zero estimate for the specific channel coefficient, as opposed to the ASVB-L algorithm, whose estimate is around zero but with higher variations in time. When the value of the true signal suddenly changes, all algorithms track the change after a few iterations. The ASVB-S and ASVB-mpL algorithms converge faster than AVSBL-L to the new signal values. In the sequel, all three algorithms track the slowly fading coefficient, with the estimates of ASVB-S and ASVB-mpL being closer to that of GARLS. 

\begin{figure}[!t]
\centering
\includegraphics[width=0.9 \linewidth]{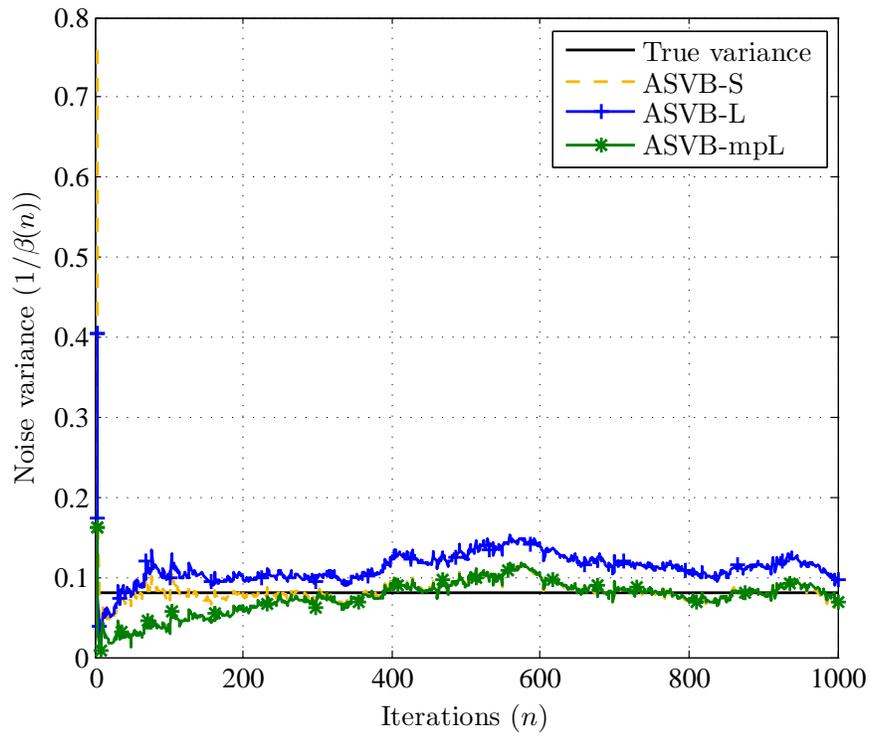}
\caption{Estimation of the noise variance in time by the proposed algorithms.}
\label{fig:beta_vs_iteration}
\end{figure}

As mentioned previously in Section \ref{sec:sbl_ad_filtering}, in contrast to all deterministic algorithms the proposed variational algorithmic framework offers the advantage of estimating not only the channel coefficients, but also the noise variance. This is a useful byproduct that can be exploited in many applications, e.g. in the area of wireless communications, where the noise variance estimate can be used when performing minimum mean square error (MMSE) channel estimation and equalization. Fig. \ref{fig:beta_vs_iteration} depicts the estimation of the noise variance offered by the Bayesian algorithms ASVB-S, ASVB-L and ASVB-mpL across time. Observe that ASVB-S and ASVB-mpL estimate accurately the true noise variance, as opposed to ASVB-L which constantly overestimates it. This is probably the reason why ASVB-L has in general inferior performance compared to ASVB-S and ASVB-mpL. 
It is worth mentioning that another useful byproduct of the variational framework is the variance of the estimates $\hat{\mathbf w}(n)$, given in (\ref{eq:sigmai}). These variances can be used to build confidence intervals for the weight estimate $\hat{\mathbf w}(n)$.

\begin{figure}[!t]
\centering
\includegraphics[width=0.9 \linewidth]{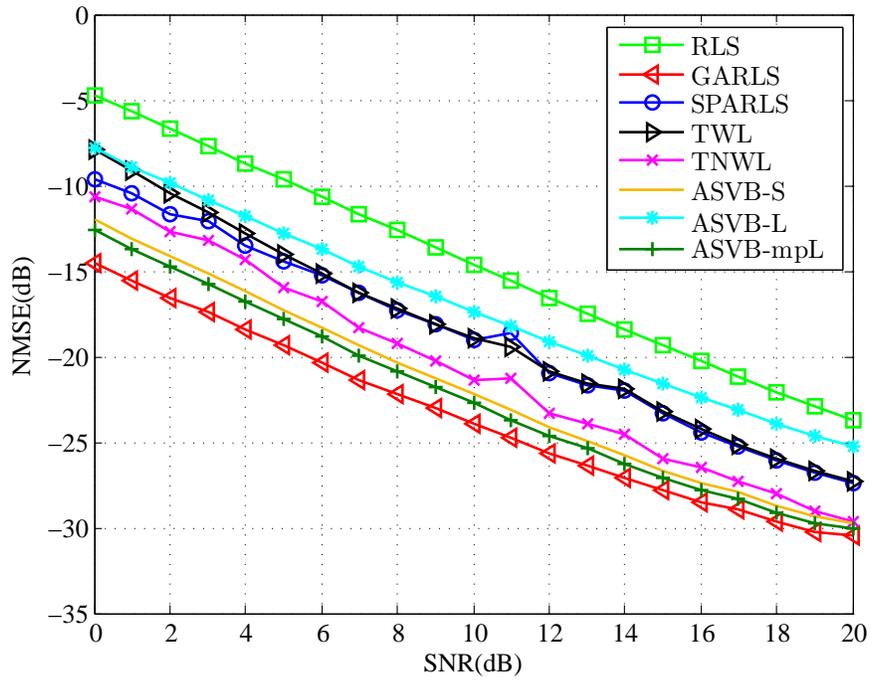}
\caption{NMSE versus SNR for all adaptive algorithms applied to the estimation of a sparse $64$-length time-varying channel with $8$ nonzero coefficients.}
\label{fig:asbl_mse_vs_snr}
\end{figure}

\begin{figure}[!t]
\centering
\includegraphics[width=0.9 \linewidth]{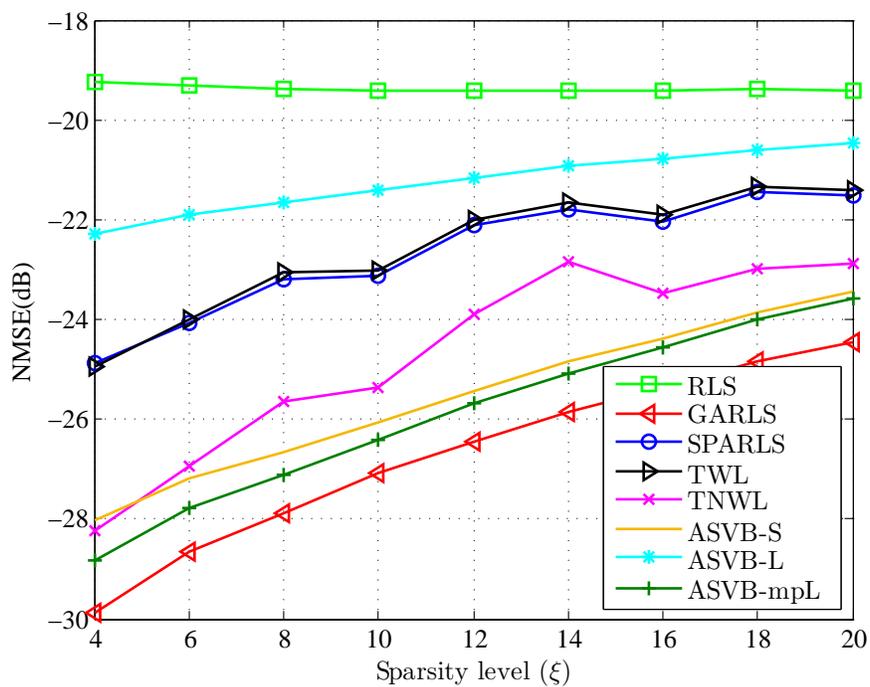}
\caption{NMSE versus the level of sparsity of the channel. The SNR is set to $15$dB. }
\label{fig:asbl_mse_vs_sparsity}
\end{figure}

The next experiments evaluate the performance of the proposed algorithms as a function of the SNR and the level of sparsity using the general settings of the first experiment. The corresponding simulation results are summarized in Figs. \ref{fig:asbl_mse_vs_snr} and \ref{fig:asbl_mse_vs_sparsity}. It can be seen in Fig. \ref{fig:asbl_mse_vs_snr} that both ASVB-S and ASVB-mpL outperform all deterministic algorithms for all SNR levels. Specifically, ASVB-mpL achieves an NMSE improvement in all SNR levels of approximately $1$dB over TNWL and $3$dB over SPARLS and TWL, as noted earlier. Moreover, in Fig. \ref{fig:asbl_mse_vs_sparsity} the curves affirm the natural increase in the NMSE of the sparsity inducing algorithms as the level of sparsity decreases. The simulation results suggest that the performance of the proposed ASVB-S and ASVB-mpL is closest to the optimal performance of GARLS, for all sparsity levels. We should also comment that only the sparsity agnostic RLS algorithm is not affected by the increase 
of the number of the channel's nonzero components.

\begin{figure}[!t]
\centering
\includegraphics[width=0.9 \linewidth]{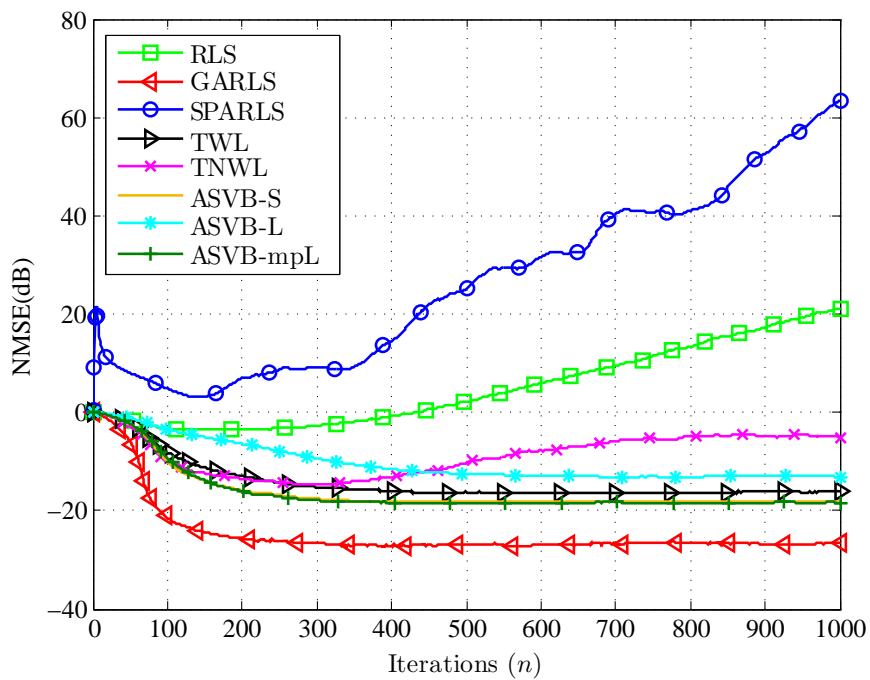}
\caption{NMSE curves of adaptive algorithms applied to the estimation of a sparse $64$-length time-varying channel with $8$ nonzero coefficients. The channel input sequence is colored using a low pass filter. The SNR is set to $15$dB.}
\label{fig:asbl_mse_vs_iteration_lowpass}
\end{figure}

As a final experiment, we test the performance of the sparse adaptive algorithms for a colored input signal. To produce a colored input sequence, a Gaussian sequence of zero mean and unit variance is lowpass filtered. For our purposes, a $5$th order Butterworth filter is used with a cut-off frequency $1/4$ the sampling rate. The remaining settings of our experiment are the same as in the first experiment. Fig. \ref{fig:asbl_mse_vs_iteration_lowpass} depicts the corresponding NMSE curves for all adaptive algorithms considered in this Section. It is clear from the figure that all algorithms' NMSE performance degrades, owning to the worse conditioning of the autocorrelation matrix ${\mathbf R}(n)$. The convergence speed of all algorithms is also slower than in Fig. \ref{fig:asbl_mse_vs_iteration}. Interestingly, RLS and SPARLS diverge. 
In addition, the poor performance of RLS has a direct impact on TNWL, since, by construction, the inverses of the RLS coefficient estimates are used to weight the $\ell_1$-norm in TNWL's cost function. Regardless, both ASVB-S and ASVB-mpL are robust, exhibiting immunity to the coloring of the input sequence.

\section{Concluding remarks}
\label{sec:conclusion}
In this paper a unifying variational Bayes framework featuring heavy-tailed priors is presented for the estimation of sparse signals and systems. Both batch and adaptive coordinate-descent type estimation algorithms with versatile sparsity promoting capabilities are described with the emphasis placed on the latter, which, to the best of our knowledge, are reported for the first time within a variational Bayesian setting. As opposed to state-of-the-art deterministic techniques, the proposed adaptive schemes are fully automated and, in addition, they naturally provide useful by-products, such as the estimate of the noise variance in time. Experimental results have shown that the new Bayesian algorithms are robust under various conditions and in general perform better than their deterministic counterparts in terms of NMSE. Extension of the proposed schemes for complex signals can be made in a straightforward manner. 
Further developments concerning analytical convergence results and faster versions of the algorithms  that update only the non-zero weights (support set) in each time 
iteration are currently under investigation.



\section{Derivation of the variational distribution \texorpdfstring{$q(w_i)$}{q(wi)}} 
\label{app:q_w_i} 
 
Starting from (\ref{eq:gen_qtheta}), the variational distribution $q(w_i)$ is computed as follows 
\begin{align} 
& q(w_i) = \frac{\mathrm{exp}\left[ \left \langle  \mathrm{log}p(\mathbf{y|\mathbf{w},\beta}) + \mathrm{log}p(\mathbf{w|\boldsymbol \alpha,\beta}) \right \rangle \right]}{\int \mathrm{exp}\left[ \left \langle  \mathrm{log}p(\mathbf{y|\mathbf{w},\beta}) + \mathrm{log}p(\mathbf{w|\boldsymbol \alpha,\beta}) \right \rangle \right]dw_i} \nonumber \\ 
& = \frac{\mathrm{exp} \left[ \left \langle  -\frac{\beta}{2} \| \boldsymbol \Lambda^{1/2} \mathbf{y} - \boldsymbol \Lambda^{1/2} \mathbf{X}_{\neg i} \mathbf{w}_{\neg i} - \boldsymbol \Lambda^{1/2} \mathbf{x}_i w_i \|^2 - \frac{\beta}{2} \alpha_i w_i^2 \right \rangle \right]}{\int \mathrm{exp} \left[ \left \langle  -\frac{\beta}{2} \| \boldsymbol \Lambda^{1/2} \mathbf{y} - \boldsymbol \Lambda^{1/2} \mathbf{X}_{\neg i} \mathbf{w}_{\neg i} - \boldsymbol \Lambda^{1/2} \mathbf{x}_i w_i \|^2 - \frac{\beta}{2} \alpha_i w_i^2 \right \rangle \right]dw_i} \nonumber \\
& = \frac{\mathrm{exp} \left[ \left \langle -\frac{\beta}{2} \left( \mathbf{x}_i^T \boldsymbol \Lambda \mathbf{x}_i w_i^2 - 2 \mathbf{x}_i^T \boldsymbol \Lambda \left( \mathbf{y} - \mathbf{X}_{\neg i} \mathbf{w}_{\neg i}\right)  w_i + \alpha_i w_i^2 \right) \right \rangle \right]}{\int \mathrm{exp} \left[ \left \langle -\frac{\beta}{2} \left( \mathbf{x}_i^T \boldsymbol \Lambda \mathbf{x}_i w_i^2 - 2 \mathbf{x}_i^T \boldsymbol \Lambda \left( \mathbf{y} - \mathbf{X}_{\neg i} \mathbf{w}_{\neg i}\right)  w_i + \alpha_i w_i^2 \right) \right \rangle \right] dw_i} \nonumber \\
& = \frac{\mathrm{exp} \left[ \left \langle -\frac{1}{2} \left( \beta \left( \mathbf{x}_i^T \boldsymbol \Lambda \mathbf{x}_i + \alpha_i \right) w_i^2  - 2 \beta  \mathbf{x}_i^T \boldsymbol \Lambda \left( \mathbf{y} - \mathbf{X}_{\neg i} \mathbf{w}_{\neg i} \right) w_i \right) \right \rangle \right]}{\int \mathrm{exp} \left[ \left \langle -\frac{1}{2} \left( \beta \left( \mathbf{x}_i^T \boldsymbol \Lambda \mathbf{x}_i + \alpha_i \right) w_i^2  - 2 \beta  \mathbf{x}_i^T \boldsymbol \Lambda \left( \mathbf{y} - \mathbf{X}_{\neg i} \mathbf{w}_{\neg i} \right) w_i \right) \right \rangle \right]dw_i} \nonumber \\
& = \frac{ \mathrm{exp} \left[ -\frac{1}{2} \left( \langle \beta \rangle \left( \mathbf{x}_i^T \boldsymbol \Lambda \mathbf{x}_i + \langle \alpha_i \rangle \right) w_i^2  - 2 \langle  \beta  \rangle  \mathbf{x}_i^T \boldsymbol \Lambda \left( \mathbf{y} - \mathbf{X}_{\neg i} \langle  \mathbf{w}_{\neg i} \rangle \right) w_i \right) \right] }{\int \mathrm{exp} \left[ -\frac{1}{2} \left( \langle \beta \rangle \left( \mathbf{x}_i^T \boldsymbol \Lambda \mathbf{x}_i + \langle \alpha_i \rangle \right) w_i^2  - 2 \langle  \beta  \rangle  \mathbf{x}_i^T \boldsymbol \Lambda \left( \mathbf{y} - \mathbf{X}_{\neg i} \langle  \mathbf{w}_{\neg i} \rangle \right) w_i \right) \right] dw_i} \nonumber \\
& = (2 \pi)^{-1/2} \sigma_i^{-1} \mathrm{exp} \left[ -\frac{1}{2} \frac{(w_i - \mu_i)^2}{\sigma_i^2}\right], 
\label{eq:q_w_i_comp} 
\end{align}  
where $\mu_i$ and $\sigma_i^2$ are given in (\ref{eq:mui}) and (\ref{eq:sigmai}) respectively.

\section{Hierarchical Laplace prior}  
\label{app:laplpr} From (\ref{eq:wprior}) and (\ref{eq:ig}) we can write, 
\begin{align} 
& p(\mathbf{w}|b,\beta) = \int p(\mathbf{w}|\boldsymbol \alpha,\beta)p(\boldsymbol \alpha|b)d\boldsymbol \alpha = \prod_{i=1}^{N}\int_{0}^{\infty} p(w_i|\alpha_i,\beta)p(\alpha_i|b)d\alpha_i \nonumber \\ 
&= (2\pi)^{-\frac{N}{2}}\beta^{\frac{N}{2}}\left(\frac{b}{2}\right)^N \prod_{i=1}^{N}\int_{0}^{\infty}\alpha_i^{-\frac{3}{2}}\mathrm{exp}\left[ -\frac{1}{2} \left(  \beta w_i^2\alpha_i + \frac{b}{\alpha_i} \right)  \right]d\alpha_i 
\label{eq:lap1}
\end{align} 
From the definition of the GIG distribution ${\cal G}{\cal I}{\cal G}(\alpha_i|\beta w_i^2,b,-\frac{1}{2})$ (cf. (\ref{eq:gig})) the integral in the last equation can be computed and (\ref{eq:lap1}) is then rewritten as 
\begin{align}
p(\mathbf{w}|b,\beta) = (2\pi)^{-\frac{N}{2}}\beta^{\frac{N}{2}}\left(\frac{b}{2}\right)^N 2^N\prod_{i=1}^{N}\frac{{\cal K}_{-1/2}(\sqrt{\beta w_i^2 b})}{\left(\frac{\beta w_i^2}{b} \right)^{-\frac{1}{4}}} 
\label{eq:lap2}
\end{align} 
In addition, 
\begin{align} 
{\cal K}_{-1/2}(x) = {\cal K}_{1/2}(x) = \sqrt{\frac{\pi}{2}}x^{-\frac{1}{2}}\mathrm{exp}\left[-x\right]. 
\label{eq:k12} 
\end{align} 
Using (\ref{eq:k12}) in (\ref{eq:lap2}) and after some straightforward simplifications, we get the multivariate Laplace distribution with parameter $\sqrt{\beta b}$, 
\begin{align} 
p(\mathbf{w}|b,\beta) = \left( \frac{\sqrt{\beta b}}{2}\right)^N\mathrm{exp}\left[-\sqrt{\beta b}\|\mathbf{w}\|_1 \right],  
\label{eq:lap3}
\end{align} 
which proves our statement.

\section{Update equation for \texorpdfstring{$\beta (n)$}{b(n)}}
\label{app:betan}
By substituting (\ref{eq:beta_exp_den}) in (\ref{eq:beta_exp}), removing $\langle \cdot \rangle$ and replacing $\boldsymbol \mu$ with $\hat{\mathbf{w}}$ yields, %
\begin{align} 
\beta = \frac{ n + N + 2\rho }{ 2\delta +  \| \boldsymbol \Lambda^{\frac{1}{2}} {\mathbf y} - \boldsymbol \Lambda^{\frac{1}{2}} \mathbf{X} \hat{\mathbf{w}}\|^2  + \sum_{i=1}^N \sigma_i^2 \mathbf{x}_i^T \boldsymbol \Lambda \mathbf{x}_i  + \sum_{i=1}^N (\hat{w}_i^2 +\sigma_i^2)\alpha_i } 
\label{eq:betapost} 
\end{align} 
Since exponentially data weighting is used, the actual time window size $n$ should be replaced by the effective time window size  $(1-\lambda)^{-1} = \sum_{j=0}^{\infty}\lambda^{j}$ and (\ref{eq:betapost}) is rewritten as 
\begin{align} 
\beta = \frac{ (1-\lambda)^{-1} + N + 2\rho }{ 2\delta +  \mathbf{y}^T \boldsymbol \Lambda \mathbf{y} -2\mathbf{z}^T\hat{\mathbf{w}} + \hat{\mathbf{w}}^T\mathbf{X}^T \boldsymbol \Lambda \mathbf{X}\hat{\mathbf{w}} + \hat{\mathbf{w}}^T \mathbf{A} \hat{\mathbf{w}} +  \sum_{i=1}^N \sigma_i^2\underbrace{(\mathbf{x}_i^T \boldsymbol \Lambda \mathbf{x}_i + \alpha_i)}_{r_{ii}}}  
\label{eq:betapost1}
\end{align} 
 or 
 \begin{align} 
 \beta = \frac{ (1-\lambda)^{-1} + N + 2\rho }{ 2\delta +  d -2\mathbf{z}^T\hat{\mathbf{w}} + \hat{\mathbf{w}}^T\mathbf{R}\hat{\mathbf{w}}  +  \boldsymbol \sigma^T\mathbf{r}}  
 \label{eq:betaexact} 
 \end{align} 
 This is an exact expression for estimating the {\it posterior} noise precision $\beta$, which can be used in the proposed algorithms. However, in order to avoid the computation of $\hat{\mathbf{w}}^T\mathbf{R}\hat{\mathbf{w}}$, which entails $N^2$ operations, we set $\hat{\mathbf{w}} = \mathbf{R}^{-1}\mathbf{z}$ in (\ref{eq:betaexact}), that is we assume that in each time iteration, $\hat{\mathbf{w}}$ attains its optimum value according to (\ref{eq:regls})\footnote{Note that this would be true if we let the Gauss-Seidel scheme iterate a few times for each $n$. On the contrary, as mentioned in Section \ref{sec:analysis}, in the proposed adaptive algorithms a {\it signle} Gauss-Seidel iteration takes place per time iteration $n$.}. Then (\ref{eq:betaexact}) is expressed as, 
 \begin{align}
 \beta = \frac{ (1-\lambda)^{-1} + N + 2\rho }{ 2\delta +  d -\mathbf{z}^T\hat{\mathbf{w}}  +  \boldsymbol \sigma^T\mathbf{r}}  
 \label{eq:betafinal} 
 \end{align} 
 Based on the update ordering of the various parameters of the algorithms in time, the respective terms in (\ref{eq:betafinal}) are expressed in terms of either $n-1$ or $n$, leading to Eq. (\ref{eq:beta_update}).




\bibliography{refs}

\begin{thebibliography}{10}
\providecommand{\url}[1]{#1}
\csname url@samestyle\endcsname
\providecommand{\newblock}{\relax}
\providecommand{\bibinfo}[2]{#2}
\providecommand{\BIBentrySTDinterwordspacing}{\spaceskip=0pt\relax}
\providecommand{\BIBentryALTinterwordstretchfactor}{4}
\providecommand{\BIBentryALTinterwordspacing}{\spaceskip=\fontdimen2\font plus
\BIBentryALTinterwordstretchfactor\fontdimen3\font minus
  \fontdimen4\font\relax}
\providecommand{\BIBforeignlanguage}[2]{{%
\expandafter\ifx\csname l@#1\endcsname\relax
\typeout{** WARNING: IEEEtran.bst: No hyphenation pattern has been}%
\typeout{** loaded for the language `#1'. Using the pattern for}%
\typeout{** the default language instead.}%
\else
\language=\csname l@#1\endcsname
\fi
#2}}
\providecommand{\BIBdecl}{\relax}
\BIBdecl

\bibitem{2001_haykin}
S.~O. Haykin, \emph{Adaptive Filter Theory}, 4th~ed.\hskip 1em plus 0.5em minus
  0.4em\relax Springer, 2002.

\bibitem{2008_sayed}
A.~H. Sayed, \emph{Adaptive Filters}.\hskip 1em plus 0.5em minus 0.4em\relax
  Wiley-IEEE Press, 2008.

\bibitem{2008_candes}
E.~Candes and M.~Wakin, ``An introduction to compressive sampling,''
  \emph{Signal Processing Magazine, IEEE}, vol.~25, no.~2, pp. 21--30, 2008.

\bibitem{2006_donoho}
D.~Donoho, ``Compressed sensing,'' \emph{Information Theory, IEEE Transactions
  on}, vol.~52, no.~4, pp. 1289--1306, 2006.

\bibitem{1996_tibshirani}
R.~Tibshirani, ``Regression shrinkage and selection via the {L}asso,''
  \emph{Journal of the Royal Statistical Society}, vol.~58, no.~1, pp.
  267--288, 1996.

\bibitem{2009_chen}
Y.~Chen, Y.~Gu, and A.~Hero, ``Sparse {LMS} for system identification,'' in
  \emph{ICASSP}, April 2009, pp. 3125 --3128.

\bibitem{2010_angelosante}
D.~Angelosante, J.~Bazerque, and G.~Giannakis, ``Online adaptive estimation of
  sparse signals: Where {RLS} meets the $\ell_1$-norm,'' \emph{Signal
  Processing, IEEE Transactions on}, vol.~58, pp. 3436 --3447, July 2010.

\bibitem{2011_eksioglu}
E.~Eksioglu and A.~Tanc, ``{RLS} algorithm with convex regularization,''
  \emph{Signal Processing Letters, IEEE}, vol.~18, no.~8, pp. 470--473, 2011.

\bibitem{2010_babadi}
B.~Babadi, N.~Kalouptsidis, and V.~Tarokh, ``{SPARLS}: The sparse {RLS}
  algorithm,'' \emph{Signal Processing, IEEE Transactions on}, vol.~58, pp.
  4013 --4025, Aug. 2010.

\bibitem{2011_kopsinis}
Y.~Kopsinis, K.~Slavakis, and S.~Theodoridis, ``Online sparse system
  identification and signal reconstruction using projections onto weighted
  $\ell_{1}$ balls,'' \emph{Signal Processing, IEEE Transactions on}, vol.~59,
  no.~3, pp. 936 --952, Mar 2011.

\bibitem{2010_mileounis}
G.~Mileounis, B.~Babadi, N.~Kalouptsidis, and V.~Tarokh, ``An adaptive greedy
  algorithm with application to nonlinear communications,'' \emph{Signal
  Processing, IEEE Transactions on}, vol.~58, pp. 2998 --3007, June 2010.

\bibitem{1999_jordan}
M.~I. Jordan, Z.~Ghahramani, T.~S. Jaakkola, and L.~K. Saul, ``An introduction
  to variational methods for graphical models,'' \emph{Machine Learning},
  vol.~37, pp. 183--233, Jan 1999.

\bibitem{2000_jaakkola}
T.~S. Jaakkola and M.~I. Jordan, ``Bayesian parameter estimation via
  variational methods,'' \emph{Statistics and Computing}, vol.~10, pp. 25--37,
  Jan. 2000.

\bibitem{2006_bishop}
C.~M. Bishop, \emph{Pattern Recognition and Machine Learning (Information
  Science and Statistics)}.\hskip 1em plus 0.5em minus 0.4em\relax Secaucus,
  NJ, USA: Springer-Verlag New York, Inc., 2006.

\bibitem{2013_themelis}
K.~E. Themelis, A.~A. Rontogiannis, and K.~Koutroumbas, ``Variational
  {B}ayesian sparse adaptive filtering using a {G}auss-{S}eidel recursive
  approach,'' in \emph{Signal Processing Conference (EUSIPCO), 2013 Proceedings
  of the 21th European}, Sep. 2013.

\bibitem{2003_koeppl}
H.~Koeppl, G.~Kubin, and G.~Paoli, ``Bayesian methods for sparse {RLS} adaptive
  filters,'' in \emph{Thirty-Seventh IEEE Asilomar Conference on Signals,
  Systems and Computers}, vol.~2.\hskip 1em plus 0.5em minus 0.4em\relax IEEE,
  2003, pp. 1273--1278.

\bibitem{2011_bioucas-dias}
J.~Bioucas-Dias, ``Bayesian wavelet-based image deconvolution: a {GEM}
  algorithm exploiting a class of heavy-tailed priors,'' \emph{Image
  Processing, IEEE Transactions on}, vol.~15, no.~4, pp. 937--951, 2006.

\bibitem{2001_girolami}
M.~Girolami, ``A variational method for learning sparse and overcomplete
  representations,'' \emph{Neural Comput.}, vol.~13, no.~11, pp. 2517--2532,
  Nov 2001.

\bibitem{2003_figueiredo}
M.~A.~T. Figueiredo and R.~Nowak, ``An {EM} algorithm for wavelet-based image
  restoration,'' \emph{Image Processing, IEEE Transactions on}, vol.~12, no.~8,
  pp. 906--916, 2003.

\bibitem{2008_park}
T.~Park and C.~George, ``The {B}ayesian {L}asso,'' \emph{Journal of the
  American Statistical Association}, vol. 103, no. 482, pp. 681--686, Jun.
  2008.

\bibitem{2012_zhang}
Z.~Zhang, S.~Wang, D.~Liu, and M.~I. Jordan, ``{EP-GIG} priors and applications
  in {B}ayesian sparse learning,'' \emph{J. Mach. Learn. Res.}, vol.~13, no.~1,
  pp. 2031--2061, Jun 2012.

\bibitem{2001_tipping}
M.~E. Tipping, ``Sparse {B}ayesian learning and the relevance vector machine,''
  \emph{J. Mach. Learn. Res.}, vol.~1, pp. 211--244, 2001.

\bibitem{2000_attias}
H.~Attias, ``A variational {B}ayesian framework for graphical models,'' in
  \emph{Advances in Neural Information Processing Systems}, vol.~12.\hskip 1em
  plus 0.5em minus 0.4em\relax MIT Press, 2000, pp. 209--215.

\bibitem{2008_tzikas}
D.~Tzikas, C.~Likas, and N.~Galatsanos, ``The variational approximation for
  {B}ayesian inference,'' \emph{Signal Processing Magazine, IEEE}, vol.~25,
  no.~6, pp. 131--146, 2008.

\bibitem{2003_beal}
\BIBentryALTinterwordspacing
M.~J. Beal, ``Variational algorithms for approximate {B}ayesian inference,''
  Ph.D. dissertation, Gatsby Computational Neuroscience Unit, University
  College London, 2003. [Online]. Available:
  \url{http://www.cse.buffalo.edu/faculty/mbeal/thesis/index.html}
\BIBentrySTDinterwordspacing

\bibitem{1987_peterson}
C.~Peterson and J.~Anderson, ``{A mean field theory learning algorithm for
  neural networks},'' \emph{Complex systems}, vol.~1, pp. 995--1019, 1987.

\bibitem{2011_shutin}
D.~Shutin, T.~Buchgraber, S.~Kulkarni, and H.~Poor, ``Fast variational sparse
  {B}ayesian learning with automatic relevance determination for superimposed
  signals,'' \emph{Signal Processing, IEEE Transactions on}, vol.~59, no.~12,
  pp. 6257--6261, 2011.

\bibitem{2010_babacan}
S.~Babacan, R.~Molina, and A.~Katsaggelos, ``Bayesian compressive sensing using
  {L}aplace priors,'' \emph{Image Processing, IEEE Transactions on}, vol.~19,
  no.~1, pp. 53--63, 2010.

\bibitem{2010_griffin}
J.~E. Griffin and P.~J. Brown, ``{Inference with normal-gamma prior
  distributions in regression problems},'' \emph{Bayesian Analysis}, vol.~5,
  no.~1, pp. 171--188, 2010.

\bibitem{2012_themelis}
K.~E. Themelis, A.~A. Rontogiannis, and K.~D. Koutroumbas, ``A novel
  hierarchical {B}ayesian approach for sparse semisupervised hyperspectral
  unmixing,'' \emph{Signal Processing, IEEE Transactions on}, vol.~60, no.~2,
  pp. 585 --599, Feb. 2012.

\bibitem{2006_zou}
H.~Zou, ``The adaptive lasso and its oracle properties,'' \emph{Journal of the
  American Statistical Association}, vol. 101, no. 476, pp. 1418--1429, Dec.
  2006.

\bibitem{2012_themelis1}
A.~A. Rontogiannis, K.~E. Themelis, and K.~Koutroumbas, ``A fast algorithm for
  the {B}ayesian adaptive lasso,'' in \emph{Signal Processing Conference
  (EUSIPCO), 2012 Proceedings of the 20th European}, Aug. 2012, pp. 974--978.

\bibitem{1996_golub}
G.~Golub and C.~Van~Loan, \emph{Matrix Computations}.\hskip 1em plus 0.5em
  minus 0.4em\relax Johns Hopkins University Press, 1996.

\bibitem{1999_xu}
X.~G-F., T.~Bose, W.~Kober, and J.~Thomas, ``A fast adaptive algorithm for
  image restoration,'' \emph{Circuits and Systems I: Fundamental Theory and
  Applications, IEEE Transactions on}, vol.~46, no.~1, pp. 216--220, 1999.

\bibitem{2004_bose}
T.~Bose, \emph{Digital signal and image processing}.\hskip 1em plus 0.5em minus
  0.4em\relax J. Wiley, 2004.

\bibitem{1998_xu}
X.~G-F. and T.~Bose, ``Analysis of the {E}uclidean direction set adaptive
  algorithm,'' in \emph{Acoustics, Speech and Signal Processing, 1998.
  Proceedings of the 1998 IEEE International Conference on}, vol.~3, 1998, pp.
  1689--1692.

\bibitem{1999_bertsekas}
D.~Bertsekas, \emph{{Nonlinear Programming}}, 2nd~ed.\hskip 1em plus 0.5em
  minus 0.4em\relax Athena Scientific, Sep. 1999.

\bibitem{1994_jakes}
W.~C. Jakes and D.~C. Cox, Eds., \emph{Microwave Mobile Communications}.\hskip
  1em plus 0.5em minus 0.4em\relax Wiley-IEEE Press, 1994.

\end{thebibliography}



\end{document}